\documentclass{article}

\PassOptionsToPackage{numbers, compress}{natbib}
\usepackage[preprint]{neurips_2026}

\usepackage[utf8]{inputenc} 
\usepackage[T1]{fontenc}    
\usepackage{hyperref}       
\usepackage{url}            
\usepackage{booktabs}       
\usepackage{amsfonts}       
\usepackage{nicefrac}       
\usepackage{microtype}      
\usepackage{xcolor}         

\usepackage{amsthm}
\usepackage{amsmath}
\newtheorem{theorem}{Theorem}
\newtheorem{lemma}[theorem]{Lemma}

\usepackage{subcaption}
\usepackage{graphicx}
\usepackage{tabularx}
\usepackage{amssymb}
\usepackage{multirow}
\newcolumntype{C}{>{\centering\arraybackslash}X}
\title{TabK: Amortized Bayesian Estimation of the Number of Clusters in Tabular Data}

\author{%
  \textbf{Mohammadreza Bakhtyari}\textsuperscript{1}\thanks{Corresponding author.}\quad
  \textbf{Bogdan Mazoure}\textsuperscript{2}\quad
  \textbf{Renato Cordeiro de Amorim}\textsuperscript{3}\\
  \textbf{Guillaume Rabusseau}\textsuperscript{2,4}\quad
  \textbf{Vladimir Makarenkov}\textsuperscript{1,2}\\[4pt]
  \textsuperscript{1}Université du Québec à Montréal, Canada\quad
  \textsuperscript{2}Mila, Canada\\
  \textsuperscript{3}University of Essex, UK\quad
  \textsuperscript{4}DIRO, Université de Montréal, Canada\\[4pt]
  \texttt{bakhtyari.mohammadreza@courrier.uqam.ca},
}

\begin{document}

\maketitle

\begin{abstract}
Estimating the number of clusters, $k$, is a fundamental prerequisite for most clustering algorithms. Traditionally, this estimation has relied on computationally expensive grid searches over internal cluster validity indices (CVIs) or adaptive clustering algorithms, both constrained by specific statistical assumptions. We introduce TabK, a zero-shot, permutation-invariant transformer architecture that frames the estimation of $k$ as an amortized Bayesian inference problem for tabular data. TabK leverages a novel Quantile Feature Encoder (QFE) to extract position-free, content-derived representations from empirical marginal distributions. By learning a generalized structural bias over a diverse synthetic prior, TabK estimates the number of clusters in a single forward pass per ensemble member. Evaluated across 50 diverse real-world benchmarks, TabK outperforms all competing methods, achieving a mean absolute error (MAE) value of 1.04 compared to 2.14 for the closest competitor. TabK also achieves the lowest average inference time, producing estimates in less than one second on average and running $2\times$ faster than TabClustPFN and $30\times$ faster than traditional CVIs such as Silhouette and Calinski--Harabasz. Code is available at \href{https://github.com/mrbakhtyari/TabK}{\nolinkurl{github.com/mrbakhtyari/TabK}}.
\end{abstract}

\section{Introduction}\label{sec:intro}
The performance of most clustering algorithms hinges on a critical prerequisite: the number of clusters, $k$, which governs partition quality. This dependency spans partition-based $k$-means \cite{lloyd1982kmeans}, graph-based spectral clustering \cite{ng2001spectral}, and probabilistic Gaussian Mixture Models (GMMs) \cite{mclachlan1988mixture}, and extends to modern deep clustering methods, including Deep Embedded Clustering (DEC) \cite{xieb2016dec}, Deep Clustering Network (DCN) \cite{yang2017dcn}, and Contrastive Clustering (CC) \cite{li2021cc}, all of which enforce an inductive bias based on a predefined $k$. In practice, however, $k$ is rarely known in advance. In domains such as customer segmentation \cite{punj1983cluster}, patient stratification \cite{wang2014similarity}, or cell-type discovery \cite{kiselev2017sc3}, the number of clusters is itself a quantity of interest. An incorrect choice of $k$ can merge distinct clusters or fragment coherent ones, thereby distorting downstream analyses.

Conventionally, determining the number of clusters in a given tabular dataset involves a validation over a discrete search space via geometric heuristics \cite{thorndike1953elbow}, internal cluster validity indices (CVIs) \cite{calinski1974ch, rousseeuw1987silhouettes}, statistically grounded null-reference comparisons \cite{tibshirani2001gapstat}, or adaptive clustering algorithms \cite{pelleg2000xmeans}. However, these approaches incur substantial computational overhead, rely on specific statistical assumptions, and introduce an inductive bias toward a predefined clustering objective. While recent deep learning approaches attempt to modernize this pipeline by integrating the estimation of $k$ directly into representation learning via joint gradient optimization \cite{ronen2022deepdpm}, they typically require dataset-specific optimization. This paradigm is particularly challenging for tabular data, which lacks shared spatial or sequential patterns common in image and text domains. The necessity for per-dataset retraining to process heterogeneous tabular data incurs high computational costs, hindering practical deployment of deep learning models in this domain \cite{grinsztajn2022trees}.

Despite these challenges, the emergence of Prior-Data Fitted Networks (PFNs) \cite{hollmann2023tabpfn} has recently redefined tabular deep learning. By employing the in-context learning capabilities of transformers, PFNs amortize Bayesian inference into a single forward pass, shifting the learning paradigm from dataset-specific gradient optimization to optimization over a diverse generative prior. While models like ZEUS \cite{marszaek2025zeus} provide zero-shot clustering on tabular datasets, they still require a predefined $k$. Subsequent architectures, such as TabClustPFN \cite{zhao2026tabclustpfn}, attempt to bridge this gap by jointly solving zero-shot clustering and cardinality inference problems. However, constraining a single model to infer global dataset structure and perform local cluster assignments tends to limit its capacity and tightly couples the resulting $k$-estimation task to a model-specific embedding space. This limitation motivates a fundamentally different design principle: treating the estimation of $k$ as a structural inference problem, which is solved independently of the downstream clustering algorithm.

This separation offers two practical advantages. First, it avoids grid search over candidate values of $k$, as is often done within clustering pipelines. Second, it allows the same predicted $k$ to be reused with different clustering algorithms. To this end, we introduce TabK, a transformer-based double-invariant architecture---ensuring permutation invariance to both objects and features---designed for zero-shot cluster cardinality estimation. Instead of relying on isolated cell-value embeddings, TabK employs a Quantile Feature Encoder (QFE) to extract learned quantile representations for each feature, explicitly encoding its global distributional statistics. Following this feature-level encoding, the model contextualizes embeddings via cross-column attention before compressing the resulting feature matrix into a dataset-level representation via inter-row attention and dynamic pooling.

Finally, acknowledging that the number of clusters follows an ordinal structure, TabK relies on Deep Label Distribution Learning (DLDL) \cite{gao2017dldl} to model a soft probability distribution over candidate $k$ values, respecting their ordinal relationships. By isolating the estimation of $k$, TabK operates as a lightweight, plug-and-play precursor that integrates seamlessly into existing clustering pipelines. Evaluating such an estimator requires a reference value of $k$. Following standard practice in the clustering literature \cite{arbelaitz2013extensive, ronen2022deepdpm, marszaek2025zeus, zhao2026tabclustpfn,bakhtyari2026ClustRecNet}, throughout the paper, we use the annotated class cardinality $k^*$ of labeled benchmarks as a proxy for the ground-truth number of clusters. In summary, our primary contributions are as follows:

\begin{itemize}
    \item We propose TabK, a transformer-based architecture that reframes the estimation of $k$ in tabular datasets from a manual, iterative, algorithm-dependent heuristic to a search-free zero-shot inference task.
    \item We introduce a quantile-based feature encoder that summarizes column-wise marginal distributions in a permutation-invariant manner.
    \item We frame estimation of the number of clusters as an ordinal-aware distribution learning task using DLDL, leveraging soft Gaussian targets and a Kullback–Leibler (KL) divergence loss to encode the ordinal relationships among candidate values of $k$.
    \item Across 50 real-world benchmarks, TabK outperforms all 17 competing methods for estimating $k$, achieving a mean absolute error (MAE) of 1.04, compared to 2.14 for the closest competitor and to 2.24 for TabClustPFN. TabK also yields the highest average Adjusted Rand Index (ARI) when its estimates of $k$ are supplied to $k$-means. Moreover, it runs more than $2\times$ faster than TabClustPFN and $30\times$ faster than classical CVIs.
\end{itemize}

\section{Related work}\label{sec:related_work}

\paragraph{Classical estimation of the number of clusters.}
Estimating the number of clusters, $k$, has historically relied on geometric heuristics and internal CVIs, both typically evaluated over a grid search. Prominent examples include the Elbow method \cite{thorndike1953elbow}, which tracks the marginal reduction of within-cluster variance, alongside established CVIs such as the Silhouette \cite{rousseeuw1987silhouettes}, Davies--Bouldin \cite{davies1979db}, Dunn \cite{dunn1974dunn}, and Calinski--Harabasz \cite{calinski1974ch} indices. To provide a more rigorous statistical grounding, the Gap Statistic \cite{tibshirani2001gapstat} formalizes structural inference by comparing observed data dispersion against a suitable null reference distribution. However, these estimation techniques rely on computationally expensive iterative execution and geometric assumptions that fail to capture heterogeneous feature spaces typical of real-world tabular data. Furthermore, because heuristic evaluation requires instantiating a specific clustering algorithm at each step of the grid search, the estimated $k$ is biased by the geometric assumptions of the chosen algorithm rather than reflecting the intrinsic structure of the data.

\paragraph{Joint estimation of the number of clusters.} To bypass external grid searches, several algorithms integrate $k$-estimation directly into the clustering objective. For instance, X-Means \cite{pelleg2000xmeans} recursively splits $k$-means clusters based on the Bayesian Information Criterion (BIC), while subsequent approaches such as Intelligent Minkowski $k$-means (IMWk) \cite{deamorim2015imwk} and Cluster Number Assisted $k$-means (CNAK) \cite{saha2021cnak} incorporate adaptive feature weighting and random data sampling. In the deep learning domain, dynamic clustering architectures have emerged to estimate the number of clusters internally. DeepDPM \cite{ronen2022deepdpm} frames the problem as a Dirichlet Process Mixture, allowing the network to split or merge clusters during training. Similarly, UNSEEN \cite{leiber2024unseen} estimates $k$ by progressively removing clusters during training, supported by a nearest-neighbor regularization loss. While these deep clustering methods eliminate the need for a grid search, they are constrained by iterative, dataset-specific gradient optimization, which is computationally expensive for heterogeneous tabular data that lacks shared inductive biases of image or text modalities.

\paragraph{Prior-data fitted networks and tabular data.} Although models like TabPFN \cite{hollmann2023tabpfn} have achieved state-of-the-art performance in supervised tabular classification, extending this paradigm to unsupervised domains remains an open challenge. In the clustering landscape, ZEUS \cite{marszaek2025zeus} successfully generates zero-shot tabular embeddings but requires an externally provided $k$ for clustering. Most closely related to our work is TabClustPFN \cite{zhao2026tabclustpfn}, which jointly performs clustering and $k$-estimation. However, its $k$-estimation is architecturally subordinate to clustering: it must compute a soft-assignment Gram matrix ($G_k = P_k^\top P_k$) for every candidate $k \in \{2, \dots, k_{\max}\}$. This design incurs $k_{\max}{-}1$ redundant decoder passes whose outputs are subsequently discarded during inference and couples $k$-estimation to the quality of the assignment logits. Rather than deriving $k$ as an artifact of cluster assignment, our work treats structural inference as a decoupled modular precursor that estimates a distribution over $k$ in a single forward pass. This decoupling frees the architecture from the need to balance local assignment accuracy against global structural inference and yields a plug-and-play module compatible with any $k$-dependent clustering algorithm.

\section{Methodology}\label{sec:method}
\begin{figure}
    \centering
    \includegraphics[width=\linewidth]{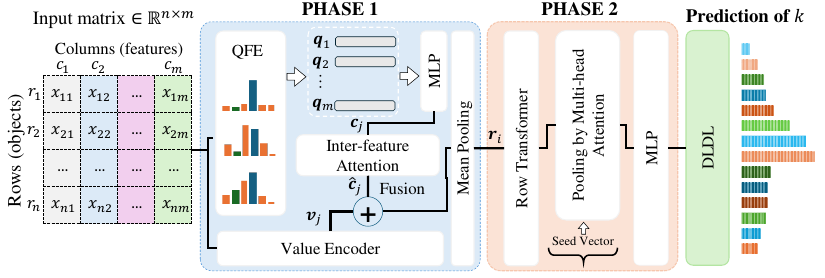}
    \caption{Overview of the TabK architecture. The model processes the input matrix $\mathbf{X}$ via a double-invariant transformer and a Quantile Feature Encoder (QFE) for zero-shot $k$ estimation.}
    \label{fig:architecture}
\end{figure}

The number of clusters, $k$, can be viewed primarily as a global property of the data distribution, rather than an instance-level quantity derived from individual object assignments to clusters. While joint-estimation models treat $k$ as an artifact of local object-to-cluster mappings and classical heuristics rely on exhaustive evaluations across varying $k$, TabK reframes the objective. Rather than predicting per-object cluster assignments, TabK operates directly on the full dataset matrix $\mathbf{X} \in \mathbb{R}^{n \times m}$ to learn a global mapping that approximates $p(k \mid \mathbf{X})$, where $n$ is the number of objects and $m$ is the number of features. To faithfully model this posterior distribution, the network must respect the permutation symmetries inherent to tabular data while bypassing the computational bottleneck of iterative local assignments. We introduce a double-invariant transformer architecture trained over a diverse generative prior with an ordinal-aware predictive head tailored to predict $k$ in tabular data. TabK first summarizes each feature using a quantile feature encoder, then models interactions among feature representations through cross-column self-attention, and finally aggregates row embeddings into a dataset-level representation via inter-row attention and dynamic pooling. This design preserves permutation invariance over both rows and columns, avoiding the introduction of arbitrary positional information. An overview of the proposed architecture is presented in Figure \ref{fig:architecture}.

\subsection{Double-invariant architecture}\label{subsec:architecture}
A core challenge in permutation-invariant tabular processing is distinguishing features without relying on explicit positional indices. While networks like TabPFN \cite{hollmann2023tabpfn} process rows as permutation-invariant sets, they map features using fixed linear projections. Consequently, they lack theoretical invariance to column permutations, and only achieve approximate column-permutation invariance empirically. Existing methods often treat individual cells as independent tokens \cite{hollmann2025tabpfnv2,lorch2022amortized}, incurring an $\mathcal{O}(nm^2)$ attention time complexity for feature-wise interactions. Alternatively, projecting the inputs into a fixed-dimensional latent space before attention reduces this complexity to $\mathcal{O}(m^2)$, but may prematurely compress the space, discarding the fine-grained marginal distributions essential for structural separability. To resolve this, TabK employs the empirical quantile function of each feature as a content-derived representation. This statistical signature provides an initial feature representation that, when combined with subsequent cross-column attention, enables the model to differentiate features based entirely on their global properties rather than arbitrary spatial coordinates. Furthermore, by isolating feature-wise interactions from row-wise dimensionality, we establish a \textit{set-of-sets} architecture. This architecture decomposes the model into two sequential phases: feature identity and row compression (Section \ref{subsub:phase1}), followed by dataset-level structural aggregation (Section \ref{subsubsec:phase2}).

\subsubsection{Phase 1: feature identity and row compression} \label{subsub:phase1}
The main objective of this phase is to transform the input matrix $\mathbf{X}$ into a sequence of row embeddings $\mathbf{r}_i \in \mathbb{R}^d$ through a process that is invariant to the input column order, capturing global feature context without combining the row and column dimensions.

\paragraph{Component 1: quantile feature encoder.} To provide an initial structural signature for features without positional indices, we encode the statistics of each column $j$ via a Quantile Feature Encoder (QFE). For a valid column vector $\mathbf{x}_{\cdot j}$, we extract $b$ evenly spaced quantiles from its empirical cumulative distribution function (CDF):
\begin{equation}
    \mathbf{q}_j = \left[ Q_{0}(\mathbf{x}_{\cdot j}), \, Q_{\frac{1}{b-1}}(\mathbf{x}_{\cdot j}), \, \ldots, \, Q_{1}(\mathbf{x}_{\cdot j}) \right] \in \mathbb{R}^b.
\end{equation}
Unlike basic moments, quantiles are order statistics that robustly capture distribution shape while remaining invariant to row permutations, with a fixed dimension regardless of the total number of objects $n$. The QFE natively captures distributional characteristics, allowing the network to leverage data artifacts, such as median imputation and categorical sparsity in one-hot encoding, as structural signals (for more details, see Appendix \ref{app:preprocessing}). Furthermore, evaluating quantiles at fixed probability levels provides a common representation of marginal distributions across different dataset sizes, which facilitates extrapolation beyond the size range of the synthetic prior (see Appendix \ref{app:scale_external}). To integrate this robust statistical profile into the network's latent space, the representation $\mathbf{q}_j$ is projected into the model dimension $d$ via a two-layer Multi-Layer Perceptron (MLP), producing the initial column identity vector \(\mathbf{c}_j\).

\paragraph{Component 2: inter-feature attention.} While marginal distributions establish an initial feature identity, each identity is computed from a single column. To contextualize each identity within the set of features, we apply a transformer encoder exclusively over the set of column identities. By running self-attention over the matrix in $\mathbb{R}^{m \times d}$ rather than the full cell representation in $\mathbb{R}^{n \times m \times d}$, we compute column identities $\hat{\mathbf{c}}_j$ at $\mathcal{O}(m^2)$, allowing features to contextualize their structural roles.

\paragraph{Component 3: value-identity fusion and row aggregation.} An isolated scalar cell-value embedding is inherently context-agnostic; it cannot ascertain whether a specific value represents an extreme outlier or a median baseline. TabK supplies this context through additive contextual fusion with the refined column identity. We project the raw scalar cell value into the latent space using learned parameters $\mathbf{w}_v$ and $\mathbf{b}_v \in \mathbb{R}^d$, and fuse it with its context-aware feature identity:
\begin{equation}
\mathbf{e}_{ij} = \underbrace{\mathbf{w}_v x_{ij} + \mathbf{b}_v}_{\text{value embedding}} + \underbrace{\hat{\mathbf{c}}_j}_{\text{refined column identity}}.
\end{equation}
Because $\hat{\mathbf{c}}_j$ already encodes the marginal distribution shape, the resulting cell representation $\mathbf{e}_{ij}$ is natively contextualized within its global statistical profile. To produce a row representation $\mathbf{r}_i$ that is invariant to column permutation, we aggregate the set of cell embeddings of row $i$, $\mathcal{F}_i = \{\mathbf{e}_{ij}\}_{j=1}^{m}$, using masked mean pooling, followed by a final projection via an MLP. This yields a vector $\mathbf{r}_i$ that combines pooled cell-value information with global column context.

\subsubsection{Phase 2: dataset-level structural representation} \label{subsubsec:phase2}
With each row now represented as a context-aware vector $\mathbf{r}_i \in \mathbb{R}^d$, the model transitions to capturing the global distributional structure of the dataset through inter-object interaction and dynamic pooling.

\paragraph{Component 4: row set interaction.} We process a collection of row embeddings $\{\mathbf{r}_1, \ldots, \mathbf{r}_n\}$ through a standard transformer encoder. By omitting positional encodings, this sequence modeling remains permutation-equivariant with respect to the rows. This inter-object attention is used for assessing dataset-level properties, enabling TabK to model relationships among row embeddings.

\paragraph{Component 5: pooling by multihead attention.} The final architectural stage aggregates the variable-size set of interacting rows into a global distributional summary $\mathbf{z} \in \mathbb{R}^d$. While standard mean pooling often attenuates the sparse signals critical for identifying rare clusters, we utilize Pooling by Multihead Attention (PMA) \cite{lee2019set}. PMA employs a learned, dataset-agnostic seed vector as a query to attend over the sequence of row embeddings. By selectively extracting representative row embeddings rather than uniformly averaging all instances, this dynamic pooling mechanism naturally adapts to varying dataset sizes and cluster imbalances.

\subsubsection{Formal symmetry guarantees} \label{subsubsec:proof}
\begin{theorem}[Double-permutation invariance]
\label{thm:double_inv}
Let $\mathcal{P}_t$ denote the set of $t \times t$ permutation matrices. For any tabular matrix $\mathbf{X} \in \mathbb{R}^{n \times m}$, row permutation $\mathbf{\Pi}_r \in \mathcal{P}_n$, and column permutation $\mathbf{\Pi}_c \in \mathcal{P}_m$, the TabK model $f_\theta$ satisfies:
\begin{equation}
    f_\theta(\mathbf{\Pi}_r \mathbf{X} \mathbf{\Pi}_c) = f_\theta(\mathbf{X})
\end{equation}
\end{theorem}
A comprehensive mathematical proof detailing the specific equivariance and invariance guarantees of each component, alongside our empirical evaluation confirming these symmetries, is provided in Appendix \ref{app:proof}. Furthermore, all architectural hyperparameters and complete optimization protocols are detailed in Appendix \ref{app:architecture}.

\subsection{Amortized Bayesian inference via generative priors}\label{subsec:amortized_inference}
TabK learns to generalize across structural patterns by amortizing Bayesian inference over a diverse synthetic prior. We define this prior, $p_{\text{prior}}(\mathbf{X}, k)$, as a joint distribution over tabular matrices $\mathbf{X}$ and their ground-truth numbers of clusters $k \in \{k_{\min}, \dots, k_{\max}\}$. During training, TabK seeks the optimal parameters $\theta^*$ that minimize the expected loss over datasets drawn from this distribution:
\begin{equation}
\theta^* = \arg\min_\theta \mathbb{E}_{(\mathbf{X}, k) \sim p_{\text{prior}}} \left[ \mathcal{L}(f_\theta(\mathbf{X}), k) \right].
\end{equation}
Instead of a hard one-hot target, we employ DLDL \cite{gao2017dldl}, whose target for a dataset with $k^*$ clusters is the discretized Gaussian:
\begin{equation}
 q_{k^*}(k) \propto \exp\!\left(-\frac{(k-k^*)^2}{2\sigma^2}\right), \quad \text{for } k \in \{k_{\min}, \dots, k_{\max}\},
\end{equation}
and the loss is $\mathcal{L} = \mathrm{KL}(q_{k^*} \,\|\, \hat{p})$, where $\hat{p}$ is the predicted distribution and $\sigma = 0.5$ controls the degree of label smoothing.

We train the model over a diverse mixture of synthetic topologies, including controlled Gaussians \cite{rodriguez2019clustering}, overlapping elliptical distributions \cite{zellinger2025natural}, elongated line-supported clusters \cite{fachada2023generating}, high-dimensional shells, and complex networked manifolds \cite{jahn2024densired} (as detailed in Appendix \ref{app:strategies}). Through this training procedure, TabK learns an approximation of $p(k\mid \mathbf{X})$ that reflects the ordinal distance between candidate values of $k$. For an unrestricted predictor, the expected loss is minimized by $\sum_{k'} q_{k'}(k)\,p_{\text{prior}}(k' \mid \mathbf{X})$, where the posterior under the synthetic prior is smoothed by the ordinal target kernel. In total, we generated 200{,}000 synthetic datasets (40{,}000 configurations $\times$ 5 strategies), 180{,}000 of which were used for training and 20{,}000 were held out as a test set (as detailed in Appendix \ref{app:config_sampling}). We train TabK using 5-fold cross-validation on the training split. At inference, we average the log-probability outputs of the models from these folds and normalize the resulting scores to obtain the ensemble distribution, whose \textit{argmax} gives the predicted $k$. All reported results, including latency, use this five-model ensemble (Appendix \ref{app:training_protocol}).

\section{Experiments}\label{sec:exp}
\subsection{Experimental setup}
\paragraph{Datasets and baselines.}
We evaluate TabK on a diverse representative suite of 50 real-world benchmarks sourced from OpenML \cite{bischl2021openml}. Because the vast majority of practical tabular datasets contain fewer than 10,000 objects \cite{hollmann2025tabpfnv2}, we scope our primary evaluation to datasets with up to 10,000 objects, 200 features, and ground-truth cluster counts $k^* \in [2, 15]$. Notably, this requires TabK to extrapolate to scales up to 3.25 times larger than those present in its synthetic training prior. Crucially, these benchmarks include missing values and categorical features, reflecting the heterogeneity of real-world tabular data (full dataset profiles are provided in Appendix \ref{app:data_preprocessing}). The benchmark was assembled to cover heterogeneous application domains and structural regimes: it spans 57 to 8{,}124 objects, 3 to 90 features, and $k^* \in [2, 15]$, and includes 16 datasets with categorical features and 9 with missing values; duplicate entries and datasets used to estimate the synthetic prior were excluded. All methods received the same preprocessed inputs: categorical features were one-hot encoded, missing values were median-imputed, and features were standardized.

We compare TabK against 17 established competitors, spanning the clustering literature: internal validity indices including Silhouette (Sil), Calinski--Harabasz (CH), Davies--Bouldin (DB), and Dunn; statistical frameworks such as Gap Statistic (Gap) and GMM-BIC \cite{fraley2002gmmbic}; and adaptive methods including X-Means, IMWk, and CNAK. We further compare TabK against deep learning architectures, specifically UNSEEN, and the state-of-the-art PFN-based method, TabClustPFN (TCP). We pair three self-supervised tabular encoders, VIME \cite{yoon2020vime}, SCARF \cite{bahri2022scarf}, and SubTab \cite{ucar2021subtab}, with the Silhouette and CH indices computed on their learned embeddings; in figures, each encoder is shown with its best-performing index (SCARF+CH, SubTab+Sil, and VIME+Sil) and labeled by the encoder name.

\paragraph{Evaluation protocol.}\label{p:eval_protocol}
To ensure a rigorous comparison, all classical baselines are evaluated across the same range of $k$ using official implementations and author-recommended hyperparameters. Latency benchmarks were standardized on an Intel Core i9-13900 CPU. Additionally, as demonstrated in Section \ref{subsec:calibration_scaling}, we evaluate how TabK's latency scales when executed under GPU acceleration (NVIDIA RTX 3060 Ti). Comprehensive implementation details for all baselines are provided in Appendix \ref{app:baseline}.

\paragraph{Metrics.} We assess the model's performance across four dimensions: structural error, prediction accuracy, downstream utility, and computational efficiency. Structural error is quantified via the MAE between the predicted $\bar{k}$ and ground truth $k^*$. Accuracy is reported as both the Exact Match (EM) rate, the proportion of datasets for which $\bar{k} = k^*$, and the $\pm 1$ Tolerance rate ($\tau = |\bar{k} - k^*| \leq 1$). To evaluate practical downstream utility, we report the ARI computed when supplying each model's predicted $\bar{k}$ to $k$-means and GMM. We additionally define the \textit{oracle} ARI as the clustering quality achieved when the ground-truth $k^*$ is supplied directly to $k$-means and GMM, providing a reference point. Finally, operational efficiency is measured via average wall-clock inference latency.

\subsection{Performance on real-world benchmarks} \label{subsec:main_results}
\begin{figure}[t]
    \centering
    \begin{subfigure}[b]{0.49\textwidth}
        \includegraphics[width=\linewidth]{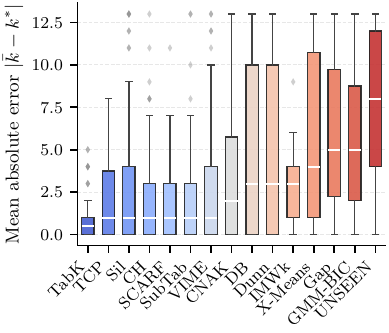}
        \caption{Error variance and stability}        \label{fig:boxplot_mae}
    \end{subfigure}
    \hfill
    \begin{subfigure}[b]{0.49\textwidth}
        \includegraphics[width=\linewidth]{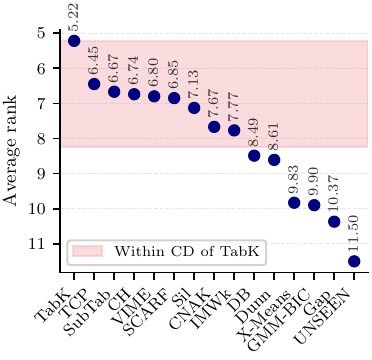}
        \caption{Average model ranking}
        \label{fig:ranking}
    \end{subfigure}
    \vfill
    \begin{subfigure}[b]{0.49\textwidth}
        \includegraphics[width=\linewidth]{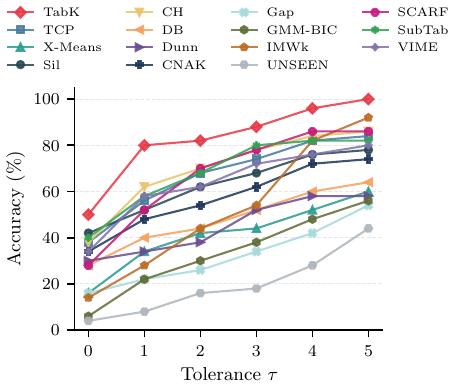}
        \caption{Error tolerance and robustness}
        \label{fig:tolerance}
    \end{subfigure}
    \hfill
    \begin{subfigure}[b]{0.49\textwidth}
        \includegraphics[width=\linewidth]{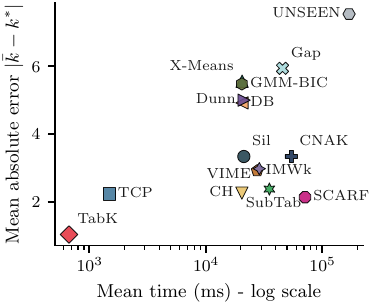}
        \caption{The Pareto frontier}
        \label{fig:pareto}
    \end{subfigure}
    \caption{Empirical evaluation on 50 OpenML real-world datasets. \textbf{(a)} MAE distributions. \textbf{(b)} Average Friedman ranks (Nemenyi test, $\alpha = 0.05$). \textbf{(c)} Cumulative accuracy at varying error tolerances ($\tau$). \textbf{(d)} Efficiency-performance Pareto frontier. TabK demonstrates the highest accuracy and average ranking, the tightest error distribution, and the lowest latency among evaluated methods.}
    \label{fig:main_results}
\end{figure}

Table \ref{tab:main_results} and Figure \ref{fig:main_results} present the aggregate performance across the 50 OpenML benchmarks. TabK exhibits the lowest median error with a low variance (see Figure \ref{fig:boxplot_mae}), in contrast to the high variance observed in both classical heuristics and modern frameworks. Moreover, TabK achieves the top average Friedman rank of 5.22 across all datasets (Figure \ref{fig:ranking}). The substantial performance margin between TabK ($\text{MAE}=1.04$; 50.0\% Exact Match) and the closest PFN-based competitor, TabClustPFN ($\text{MAE}=2.24$; 38.0\% Exact Match), provides empirical support for the proposed decoupled architectural design. To ensure this gap is not an artifact of different prior scale, we evaluated TabK within TabClustPFN's training-scale upper bounds ($n \leq 1000$, $m \leq 64$, $k \leq 10$). As detailed in Appendix \ref{app:hometurf}, TabK outperforms TabClustPFN even within these constraints. Furthermore, TabK's MAE improvements are statistically significant against each competitor (Wilcoxon test $p$-value $< 0.05$). This advantage does not stem from ensembling alone: a single ensemble member already attains an average MAE of 2.06, which is better than the results of all competing methods. Moreover, TabK's estimates are robust under irrelevant features and outliers: its MAE rises to at most 1.94, which is still lower than the MAE of every competing method on the original data (Appendix~\ref{app:robustness}).

\begin{table}[t]
\caption{Aggregate performance on 50 OpenML benchmarks. TabK achieves the lowest MAE, and the highest Exact Match (EM) and $\pm 1$ Tolerance rates. $\text{ARI}_{1}$ and $\text{ARI}_{2}$ correspond to the ARIs obtained with $k$-means and GMM, respectively. Best results are \textbf{bolded}, second-best \underline{underlined}.}
\label{tab:main_results}
\centering
\small
\setlength{\tabcolsep}{5pt}
\begin{tabular}{l c c c c c c c}
\toprule
& \multicolumn{6}{c}{\textbf{Performance metrics}} & \multicolumn{1}{c}{\textbf{Wilcoxon test}} \\
\cmidrule(lr){2-7} \cmidrule(lr){8-8}
Method & MAE $\downarrow$ & $\text{ARI}_{1}$ $\uparrow$ & $\text{ARI}_{2}$ $\uparrow$ & EM (\%) $\uparrow$ & $\tau \leq 1$ (\%) $\uparrow$ & Latency (s) $\downarrow$ & $p$-value \\
\midrule
\textbf{TabK} & \textbf{1.04} & \textbf{0.232} & \underline{0.221} & \textbf{50.0} & \textbf{80.0} & \textbf{0.67} & -- \\
SCARF+CH & \underline{2.14} & 0.218 & 0.207 & 28.0 & 52.0 & 71.3 & $1.5 \times 10^{-3}$ \\
TCP & 2.24 & 0.210 & 0.213 & 38.0 & 56.0 & \underline{1.50} & $6.5 \times 10^{-3}$ \\
CH & 2.26 & 0.207 & 0.200 & 38.0 & \underline{62.0} & 20.5 & $1.7 \times 10^{-2}$ \\
SubTab+Sil & 2.38 & 0.214 & 0.213 & 40.0 & 58.0 & 35.2 & $1.3 \times 10^{-2}$ \\
SCARF+Sil & 2.50 & 0.200 & 0.166 & 36.0 & 54.0 & 72.9 & $3.6 \times 10^{-3}$ \\
IMWk & 2.92 & 0.221 & 0.210 & 14.0 & 28.0 & 27.7 & $6.0 \times 10^{-5}$ \\
VIME+Sil & 2.98 & \underline{0.230} & \textbf{0.243} & 34.0 & 58.0 & 28.9 & $3.9 \times 10^{-3}$ \\
VIME+CH & 3.08 & 0.190 & 0.201 & \underline{42.0} & 56.0 & 27.4 & $7.1 \times 10^{-3}$ \\
Sil & 3.34 & 0.202 & 0.199 & \underline{42.0} & 52.0 & 21.3 & $2.0 \times 10^{-3}$ \\
CNAK & 3.34 & 0.203 & 0.199 & 34.0 & 48.0 & 54.2 & $1.8 \times 10^{-4}$ \\
SubTab+CH & 3.72 & 0.211 & 0.205 & 32.0 & 52.0 & 33.9 & $8.0 \times 10^{-4}$ \\
DB & 4.92 & 0.205 & 0.210 & 28.0 & 40.0 & 20.5 & $1.7 \times 10^{-5}$ \\
Dunn & 5.00 & 0.183 & 0.194 & 30.0 & 34.0 & 21.4 & $2.7 \times 10^{-6}$ \\
GMM-BIC & 5.48 & 0.188 & 0.198 & 6.00 & 22.0 & 20.4 & $1.6 \times 10^{-7}$ \\
X-Means & 5.56 & 0.228 & 0.220 & 16.0 & 34.0 & 20.5 & $4.9 \times 10^{-7}$ \\
Gap & 5.94 & 0.191 & 0.195 & 16.0 & 22.0 & 45.7 & $4.9 \times 10^{-8}$ \\
UNSEEN & 7.54 & 0.171 & 0.166 & 4.00 & 8.00 & 170 & $3.7 \times 10^{-9}$ \\
\bottomrule
\end{tabular}
\end{table}

\begin{figure}[t]
    \centering
    \includegraphics[width=0.95\linewidth]{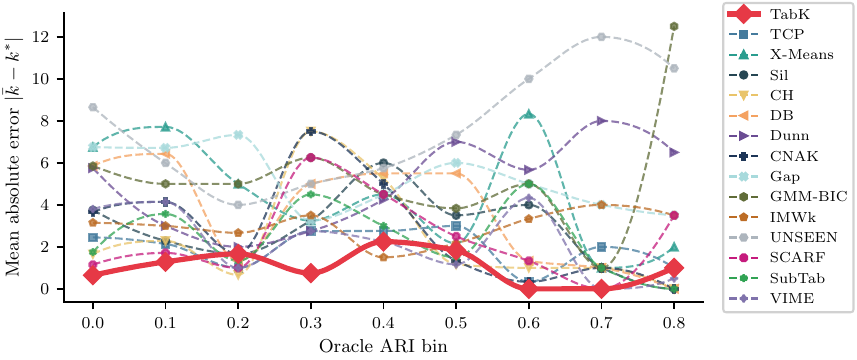}
    \caption{TabK performance across difficulty strata. MAE remains bounded across all \textit{oracle} ARIs of $k$-means, whereas evaluated competitors exhibit high variance and marked failures. Datasets are grouped by \textit{oracle} ARI into bins of width 0.1, and markers show the mean MAE per bin.}
    \label{fig:ari_strata}
\end{figure}

Beyond Exact Match, TabK exhibits strong error boundedness. TabK outperforms all evaluated methods at every tolerance threshold $\tau$ (see Figure \ref{fig:tolerance}), achieving a $\pm 1$ Tolerance rate ($\tau \leq 1$) of 80.0\%. In real-world tabular domains---where the ground-truth $k$ is frequently ambiguous and dependent on human annotation---this bounded error is highly practical. Thus, when TabK diverges from the ground truth, it degrades gracefully, confining its predictions to values close to \(k^*\). Crucially, this precise estimation of $k$ translates directly to downstream clustering. When supplying the predicted \(\bar{k}\) to standard $k$-means, TabK yields the highest ARI (0.232), recovering 92.18\% of the \textit{oracle} ARI (0.251), while maintaining a bounded MAE across difficulty strata (see Figure \ref{fig:ari_strata}). It is worth noting that this advantage is not driven by datasets whose labels are poorly aligned with geometric structure: restricted to the 20 datasets with the \textit{oracle} ARI $\geq 0.30$, TabK attains an MAE of 1.25 vs 2.05 for the closest competitor (VIME+Sil), and it ranks first at every threshold between 0.10 and 0.30. When $\bar{k}$ is supplied to a GMM, TabK yields an average ARI of 0.221, recovering 85.20\% of the \textit{oracle} ARI (0.259), and outperforms all raw-feature baselines, including algorithms explicitly optimized for Gaussian geometries such as X-Means (0.220) and GMM-BIC (0.198); only VIME+Sil attains a higher average GMM ARI (0.243), with a substantially higher MAE (2.98).

Figure \ref{fig:pareto} illustrates the Pareto frontier, revealing that TabK achieves a highly favorable efficiency-performance trade-off among all evaluated methods. By amortizing the search into one forward pass per ensemble member, TabK executes in an average of 0.67 seconds per dataset. This represents a more than 2$\times$ speedup over TabClustPFN (1.5s) and far surpasses the exhaustive iterative grid searches used by classical baselines (requiring $\geq$20s).

\subsection{Predictive uncertainty and scalability analysis}\label{subsec:calibration_scaling}
While aggregate metrics establish TabK's performance advantage, a deeper analysis against the closest PFN-based competitor, TCP, reveals improvements in predictive uncertainty and empirical scaling. Since both models output probability distributions and support GPU acceleration, this comparison highlights differences in predictive uncertainty and computational latency.

\begin{figure}[t]
    \centering
    \begin{subfigure}[b]{0.48\textwidth}
        \includegraphics[width=\linewidth]{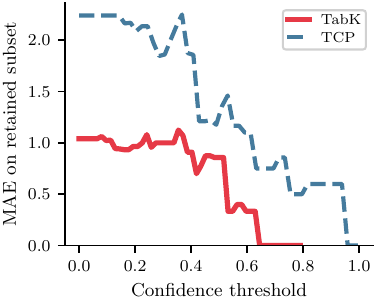}
        \caption{Risk-coverage analysis}
        \label{fig:risk_coverage}
    \end{subfigure}
    \hfill
    \begin{subfigure}[b]{0.48\textwidth}
        \includegraphics[width=\linewidth]{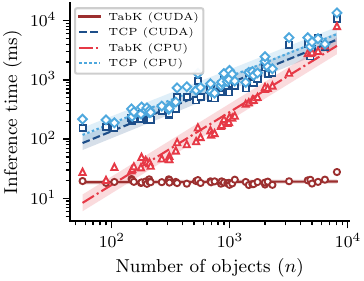}
        \caption{Empirical scaling (latency vs. size)}
        \label{fig:scale}
    \end{subfigure}
    \caption{Head-to-head analysis of TabK vs. TabClustPFN (TCP). \textbf{(a)} Risk-coverage analysis showing TabK's and TCP's confidence ranking.
    \textbf{(b)} Log-log scalability plot demonstrating TabK's order-of-magnitude speedup on GPU, with CPU execution frequently outpacing TCP on GPU.}
    \label{fig:deep_dive}
\end{figure}

\paragraph{Risk-coverage.} Because TabK frames the estimation of $k$ as a distribution-aware learning task via DLDL, the network outputs a probability mass function over the discrete candidate space of $k$. Thus, we measure confidence as the geometric mean (across the five ensemble members) of the probability assigned to the predicted value of $k$; this confidence decreases when the members disagree. To evaluate whether this confidence reflects prediction error, we perform a risk-coverage analysis (see Figure \ref{fig:risk_coverage}). TabK demonstrates reliable confidence ranking: as the confidence threshold increases, the MAE generally decreases, approaching zero error on its most confident predictions. Conversely, TabClustPFN exhibits a less monotonic risk-coverage curve, with MAE fluctuating even at higher confidence thresholds. TabK also achieves a lower negative log-likelihood (NLL) of 2.08, compared to 3.07 for TabClustPFN, confirming improved alignment between predicted distributions and observed outcomes.

\paragraph{Computational scalability.} Figure \ref{fig:scale} compares the empirical inference latency of TabK and TabClustPFN across varying dataset sizes. On GPU, TabK achieves an order-of-magnitude speedup over its main competitor. Notably, in 96\% of cases, TabK, executed on a standard CPU, remains faster than TabClustPFN, executed on a dedicated GPU,  highlighting the practical efficiency of the proposed architecture. Beyond the primary benchmark, TabK also outperforms TCP on 25 OpenML-CC18 datasets, including 10 that exceed the size of the largest synthetic training table by up to $13\times$ (Appendix~\ref{app:scale_external}).

\subsection{Ablation study}\label{sec:ablations}
To validate the proposed TabK architecture, we systematically ablated its core architectural components, predictive heads, and synthetic prior configurations. Table \ref{tab:ablations} reports the MAE results obtained for the set of 50 OpenML benchmarks. The reference TabK model, trained on the full mixture prior with a DLDL head and the complete architecture, achieves an MAE of 1.04.

\paragraph{Architectural components.} The largest architectural degradation occurs when the QFE is replaced with standard parametric moments, increasing the MAE to 1.82. While higher-order parametric moments may partially recover this performance, quantiles provide a non-parametric alternative that requires no distributional assumptions, making them more appropriate for heterogeneous tabular data. Similarly, removing cross-column attention ($\text{MAE}=1.48$) restricts the model's ability to capture interactions among column representations. Finally, replacing PMA with standard mean pooling ($\text{MAE}=1.38$) impairs the model's capacity to selectively attend to representative rows.

\paragraph{Predictive head formulations.} TabK uses DLDL to capture the ordinal relationships between candidate values of $k$. Reverting to standard categorical classification ($\text{MAE}=1.40$) or Focal Loss ($\text{MAE}=1.44$) degrades performance, as these objectives treat all misclassifications as equally incorrect, failing to impose a penalty proportional to the distance from $k^*$. Conversely, scalar regression ($\text{MAE}=2.10$) treats $k$ as a continuous variable, ignoring its discrete nature. The ordinal head suffers the most severe degradation ($\text{MAE}=2.34$). DLDL provides the most reliable objective, as its soft Gaussian targets natively bridge these gaps, respecting the discrete, ordinal hierarchy of $k$.

\paragraph{Prior volume and diversity.} Amortized Bayesian inference benefits from the scale and diversity of the generative prior. Training on only 25\% of the synthetic datasets yields an MAE of 1.72, with performance improving monotonically with prior volume and the most substantial gain observed at the full prior. Crucially, every generator contributes to generalization: removing any single generator from the mixture, while keeping the prior size fixed, increases the MAE from 1.04 to between 1.70 (without networked manifolds) and 3.56 (without elongated line-supported clusters). Thus, training on the full mixture prior yields the best performance overall, allowing TabK to generalize across different cluster geometries and adapt to variable tabular distributions.

\begin{table}[t]
  \caption{Ablation study results in terms of MAE evaluated across architectural components, predictive heads, and synthetic prior configurations. The $\Delta$ column indicates performance degradation relative to the full TabK model with an MAE of 1.04. Leave-one-generator-out removes a single generator from the prior mixture while keeping the total number of synthetic datasets fixed.}
  \label{tab:ablations}
  \centering
  \begin{tabular}{l c c @{\hspace{10pt}} l c c}
    \toprule
    \textbf{Model variation} & \textbf{MAE $\downarrow$} & \textbf{$\Delta$} & \textbf{Prior variation} & \textbf{MAE $\downarrow$} & \textbf{$\Delta$} \\
    \midrule
    \multicolumn{3}{l}{\textit{Architectural components}} & \multicolumn{3}{l}{\textit{Prior volume}} \\
    \hspace{4mm} w/o QFE & 1.82 & +0.78 & \hspace{4mm} 25\% of synthetic prior & 1.72 & +0.68 \\
    \hspace{4mm} w/o column attention & 1.48 & +0.44 & \hspace{4mm} 50\% of synthetic prior & 1.46 & +0.42 \\
    \hspace{4mm} w/o PMA & 1.38 & +0.34 & \hspace{4mm} 75\% of synthetic prior & 1.42 & +0.38 \\
    \cmidrule(r){1-3} \cmidrule(l){4-6}
    \multicolumn{3}{l}{\textit{Head formulations}} & \multicolumn{3}{l}{\textit{Leave-one-generator-out}} \\
    \hspace{4mm} Ordinal & 2.34 & +1.30 & \hspace{4mm} w/o networked \cite{jahn2024densired} & 1.70 & +0.66 \\
    \hspace{4mm} Regression & 2.10 & +1.06 & \hspace{4mm} w/o concentric shells & 1.84 & +0.80 \\
    \hspace{4mm} Focal Loss & 1.44 & +0.40 & \hspace{4mm} w/o elliptical \cite{zellinger2025natural} & 1.84 & +0.80 \\
    \hspace{4mm} Classification & 1.40 & +0.36 & \hspace{4mm} w/o Gaussians \cite{rodriguez2019clustering} & 1.86 & +0.82 \\
    & & & \hspace{4mm} w/o elongated \cite{fachada2023generating} & 3.56 & +2.52 \\
    \bottomrule
  \end{tabular}
\end{table}

\section{Conclusion}\label{sec:conclusion}
We introduced TabK, a double-invariant architecture designed to transition the estimation of the number of clusters, $k$, from a computationally expensive iterative heuristic to a rapid, zero-shot inference task. By treating $k$-estimation as an independent, global distributional objective rather than a localized object-assignment problem, TabK provides a modular, accurate, and efficient precursor for downstream clustering. Equipped with a Quantile Feature Encoder to process feature distributions and an ordinal-aware Deep Label Distribution Learning objective, TabK outperforms all evaluated methods in estimating $k$, achieving state-of-the-art performance on real-world benchmarks within the evaluated operating window ($n \leq 10{,}000$, $m \leq 200$, $k \in [2, 15]$). Furthermore, by amortizing inference over a diverse synthetic prior, TabK estimates $k$ in less than one second on average. A thorough discussion of the limitations of this work, including the generative prior constraints, computational complexity, and ground-truth ambiguity, is presented in Appendix \ref{app:limitation}.

\begin{ack}
This work was supported by grant 371537 through the partnership between the Ministère de l'Enseignement supérieur du Québec and the Fonds de recherche du Québec, and by the Natural Sciences and Engineering Research Council of Canada (NSERC) under grant 249644. The authors declare no competing interests.
\end{ack}

\bibliographystyle{plainnat}
\bibliography{references}


\appendix
\newpage

\section{Limitations} \label{app:limitation}
\paragraph{Generative prior scale and generalization.} Although amortized inference over a synthetic prior eliminates iterative grid searches, its synthetic training prior covers a finite range: up to 2,500 objects, 200 features, and 15 clusters. While this represents a significant expansion over previous PFNs (e.g., TabClustPFN limits its domain to 1,000 objects, 64 features, and 10 clusters), the prediction head remains restricted to $k \in \{2,\ldots,15\}$. Nevertheless, the evaluation in Appendix~\ref{app:scale_external} provides empirical evidence of extrapolation in both object and feature counts, with datasets reaching 96{,}320 objects and 1{,}776 features, processed without subsampling or truncation. On the 10 datasets with more cells than the largest training table (up to $13\times$ as many), TabK achieves an MAE of 1.00. We attribute this extrapolation in object count to the QFE, whose quantile-based encoding produces fixed-size representations by evaluating empirical quantiles at fixed probability levels.

\paragraph{Computational complexity of attention.} A further limitation stems from the quadratic time and memory complexity of the transformer attention mechanism across both the row and column dimensions. In practice, TabK mitigates the practical computational cost on GPUs through automatic mixed-precision (AMP) (detailed in Appendix \ref{app:training_protocol}), which contributes to the inference speedup reported in Figure \ref{fig:scale}. In addition, on the primary benchmarks exceeding the training object-count ceiling, we show in Appendix \ref{app:subsampling} that row subsampling largely preserves prediction consistency. Investigating and integrating linear attention mechanisms adapted to the tabular domain is a promising direction for extending zero-shot inference to million-row datasets.

\paragraph{Ground-truth ambiguity in evaluation.} A fundamental challenge in evaluating unsupervised algorithms is the reliance on human-annotated class labels to define the ground-truth number of clusters, $k^*$. In reality, these labels frequently encode semantic or arbitrary boundaries that do not necessarily correspond to distinct, geometrically separable clusters. Consequently, the annotations themselves often fail to provide a well-defined ground truth for the clustering task. On hard datasets---where $k$-means used with $k^*$ yields a low \textit{oracle} ARI---a low MAE does not unambiguously certify that a model has recovered the correct geometric structure. It may merely indicate that the model has learned to predict the human-annotated $k^*$, regardless of intrinsic spatial separability. As demonstrated in Figure \ref{fig:ari_strata}, TabK maintains bounded prediction error relative to the annotated $k^*$ across the evaluated real-world benchmarks. Nevertheless, practitioners should interpret standard evaluation metrics cautiously; a model predicting a geometrically coherent number of clusters that differs from $k^*$ may be more useful than one that matches the subjective annotation. We therefore recommend treating TabK's probabilistic output as a data-driven estimation of $k$ to be validated against domain knowledge.

\section{Synthetic prior generation} \label{app:synthetic_data_gen}
TabK's amortized framework requires a synthetic prior $p_{\text{prior}}(\mathbf{X}, k)$ whose geometric diversity is designed to approximate the heterogeneous real-world tabular data. We therefore construct a structured prior by decoupling two complementary design decisions. First, we sample configuration triples $(k, n, m)$ from distributions grounded in real tabular benchmarks. Second, we instantiate each configuration through five generative strategies, each targeting a distinct cluster geometry. Together, these two components ensure that TabK is exposed during training to a configuration space and a geometric diversity that jointly approximate the heterogeneous structure of real-world tabular data.

\subsection{Configuration sampling} \label{app:config_sampling}
Each synthetic dataset is parameterized by a configuration triple $(k,n,m)$, representing the number of clusters, objects, and features, respectively. To ground this prior in real-world statistics, we collected classification datasets from OpenML, filtered to those whose $(k,n,m)$ fell within our target ranges $k \in [2, 15]$, $n \in [100, 2500]$, and $m \in [2, 200]$, and removed duplicates. We used this curated corpus to choose parametric marginals for $k$, $n$, and $m$ that reflect the empirical distributional landscape of real tabular benchmarks.

We generated $N_{\text{cfg}} = 40{,}000$ configurations using Latin Hypercube Sampling (LHS) \cite{mckay1979lhs} over $[0,1)^3$. Each uniform marginal $(u_k, u_n, u_m)$ was mapped through the corresponding inverse CDF: a truncated Zipf distribution ($\alpha = 1.6$) for $k$, and log-uniform distributions for $n$ and $m$, providing stratified coverage of the specified marginal distributions.

\subsection{Generative strategies} \label{app:strategies}
\begin{figure}[t]
    \centering
    \begin{subfigure}[b]{0.19\textwidth}
        \centering
        \includegraphics[width=\textwidth]{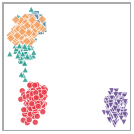}
        \caption{}
    \end{subfigure}\hfill
    \begin{subfigure}[b]{0.19\textwidth}
        \centering
        \includegraphics[width=\textwidth]{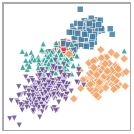}
        \caption{}
    \end{subfigure}\hfill
    \begin{subfigure}[b]{0.19\textwidth}
        \centering
        \includegraphics[width=\textwidth]{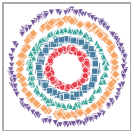}
        \caption{}
    \end{subfigure}\hfill
    \begin{subfigure}[b]{0.19\textwidth}
        \centering
        \includegraphics[width=\textwidth]{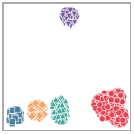}
        \caption{}
    \end{subfigure}\hfill
    \begin{subfigure}[b]{0.19\textwidth}
        \centering
        \includegraphics[width=\textwidth]{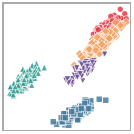}
        \caption{}
    \end{subfigure}
    \caption{Visualization of synthetic geometries used for the generative prior ($n=1000, m=2, k=5$): (a) Gaussian clusters \cite{rodriguez2019clustering}, (b) anisotropic ellipsoids \cite{zellinger2025natural}, (c) non-convex concentric shells, (d) networked manifolds \cite{jahn2024densired}, and (e) elongated line-supported clusters \cite{fachada2023generating}.}
    \label{fig:strategies}
\end{figure}

To ensure TabK learns generalized structural representations, we constructed a diverse synthetic prior using a mixture of five distinct geometric generation strategies, illustrated in Figure \ref{fig:strategies}. Strategy-specific hyperparameters were sampled independently per instance, ensuring that identical $(k,n,m)$ triples generate geometrically distinct datasets. Table \ref{tab:strategies} summarizes all hyperparameter distributions.

\begin{table}[t]
    \centering
    \small
    \renewcommand{\arraystretch}{1.1}
    \setlength{\tabcolsep}{9pt}
    \caption{Hyperparameter specifications and sampling distributions used within our synthetic data generation strategies.}
    \label{tab:strategies}
    \begin{tabular}{l l l}
        \toprule
        \textbf{Strategy} & \textbf{Parameter} & \textbf{Value ranges / distributions} \\
        \midrule
        \citet{rodriguez2019clustering}
        & $\alpha$ & Log-Uniform $[0.05, 5.0]$ \\

        \midrule
        \citet{fachada2023generating}
        & \texttt{direction}    & Random Unit Vector \\
        & \texttt{angle\_disp}  & Uniform $[0.1, 0.6]$ \\
        & \texttt{llength}     & Log-Uniform $[10.0, 60.0]$ \\
        & \texttt{cluster\_sep} & $(\texttt{llength}/\sqrt{m})\times[0.1,0.6]$ per coordinate \\
        & \texttt{llength\_disp}& $\texttt{llength} \times [0.6, 1.1]$ \\
        & \texttt{lateral\_disp}& Uniform $[0.5, 3.0]$ \\
        & \texttt{proj\_dist\_fn}& Norm (0.6), Unif (0.4) \\
        & \texttt{point\_dist\_fn}& \texttt{n} (0.8), \texttt{n-1} (0.2) \\

        \midrule
        \citet{jahn2024densired}
        & \texttt{radius}        & Log-Uniform $[0.6, 1.8]$ \\
        & \texttt{step}          & $\texttt{radius} \times [0.7, 1.6]$ \\
        & \texttt{min\_dist}      & Uniform $[0.92, 1.06]$ \\
        & \texttt{ratio\_noise}   & Uniform $[0.05, 0.35]$ \\
        & \texttt{connections}   & $\{0\} \cup [1, \min(6, \text{max\_edges})]$ \\
        & \texttt{ratio\_con}     & $\{0\} \cup \text{Unif } [0.03, 0.25]$ \\
        & \texttt{con\_min\_dist} & Uniform $[0.9, 1.05]$ \\
        & \texttt{con\_step}      & $\texttt{step} \times \text{Unif } [0.6, 1.1]$ \\
        & \texttt{dens\_factors}  & Boolean ($P=0.8$) \\
        & \texttt{momentum}      & Beta(1.5, 1.5) $\to [0.25, 0.9]$ \\
        & \texttt{distribution}  & \texttt{studentt} (0.5), \texttt{gaussian} (0.3), \texttt{paper} (0.2) \\

        \midrule
        \citet{zellinger2025natural}
        & \texttt{min\_max\_overlap} & (0.001, 0.002), (0.002, 0.25), (0.25, 0.5) \\
        & \texttt{transform\_type}   & none (0.6), distort (0.2), wrap (0.2) \\
        & \texttt{aspect\_ref}       & $\{1, 3\}$ \\
        & \texttt{aspect\_maxmin}    & $\{1, 3\}$ \\
        & \texttt{radius\_maxmin}    & $\{1, 3, 10\}$ \\
        & \texttt{imbalance\_ratio}  & $\{1, 3, 10\}$ \\

        \midrule
        Concentric Hyperspheres & \texttt{noise} & Uniform $[0.005, 0.1]$ \\
                     & \texttt{factor} & Uniform $[0.4, 0.8]$ \\
        \bottomrule
    \end{tabular}
\end{table}

The first strategy follows the method by \citet{rodriguez2019clustering} that generates Gaussian clusters via cluster-specific factor matrices with random centroid offsets, where a scaling coefficient $\alpha$ controls the degree of separation, ranging from heavily overlapping to well-separated configurations. The second strategy uses the Repliclust framework \cite{zellinger2025natural} to produce ellipsoidal clusters with anisotropic covariance, heterogeneous scales, and calibrated overlap, optionally passed through a nonlinear manifold transform. The third strategy samples objects uniformly on the $m$-dimensional unit hypersphere and arranges them into concentric shells of increasing radius, producing non-convex nested geometries that linear and centroid-based methods struggle to cluster. The fourth strategy employs the Densired generator \cite{jahn2024densired} to construct cluster cores as random-walk graphs with optional inter-cluster bridging filaments, per-cluster density scaling, and background noise, capturing some of the complex irregular geometries common in real-world data. The fifth strategy uses PyClugen \cite{fachada2023generating} to distribute objects along randomly oriented line segments in $\mathbb{R}^m$, yielding elongated cigar-shaped clusters with variable orientation and lateral scatter.

\section{Architecture} \label{app:architecture}
\subsection{Proof of double-permutation invariance}\label{app:proof}
In this section, we provide the formal proof of Theorem \ref{thm:double_inv}. We demonstrate that the TabK architecture preserves symmetry by verifying how the permutations transform during the two sequential phases of the model.

\begin{lemma}[Equivariance of transformer layers]
\label{lemma:transformer}
Let $\text{T}(\cdot)$ be a transformer encoder layer without positional encodings. For any permutation matrix $\mathbf{\Pi} \in \mathcal{P}_t$ acting on a sequence $\mathbf{Z} \in \mathbb{R}^{t \times d}$, the operation is permutation-equivariant: $\text{T}(\mathbf{\Pi}\mathbf{Z}) = \mathbf{\Pi}\text{T}(\mathbf{Z})$.
\end{lemma}
\begin{proof}
The self-attention mechanism is defined as $\text{Attn}(\mathbf{Q, K, V}) = \text{Softmax}(\frac{\mathbf{QK}^\top}{\sqrt{d}})\mathbf{V}$. Since $\mathbf{Q, K}$, and $\mathbf {V}$ are row-wise linear projections of $\mathbf{Z}$, permuting the rows of $\mathbf{Z}$ results in $\mathbf{\Pi}\mathbf{Q}, \mathbf{\Pi}\mathbf{K}$, and $\mathbf{\Pi}\mathbf{V}$, respectively. The resulting attention map $\mathbf{\Pi}\mathbf{A}\mathbf{\Pi}^\top$ is permuted both row-wise and column-wise. Multiplying this by the permuted value matrix yields $\mathbf{\Pi}(\mathbf{A}\mathbf{V})$, preserving the permutation order \cite{vaswani2017attention, lee2019set}.
\end{proof}

\paragraph{Theorem \ref{thm:double_inv}.}
Let $\mathcal{P}_t$ denote the set of $t \times t$ permutation matrices. For any tabular matrix $\mathbf{X} \in \mathbb{R}^{n \times m}$, row permutation $\mathbf{\Pi}_r \in \mathcal{P}_n$, and column permutation $\mathbf{\Pi}_c \in \mathcal{P}_m$, the TabK model $f_\theta$ satisfies:
 \begin{equation}
     f_\theta(\mathbf{\Pi}_r \mathbf{X} \mathbf{\Pi}_c) = f_\theta(\mathbf{X})
 \end{equation}
\begin{proof}
Let $\mathbf{X}$ be the input matrix transformed by row permutation $\mathbf{\Pi}_r$ and column permutation $\mathbf{\Pi}_c$. We denote the transformed input as $\mathbf{X}' = \mathbf{\Pi}_r \mathbf{X} \mathbf{\Pi}_c$.

\paragraph{Phase 1: Feature identity and row compression.}
First, we consider \textbf{Component 1 (QFE)}. The empirical quantiles $\mathbf{q}_j$ are calculated over the set of values in column $j$. Since the set $\{x_{1j}, \dots, x_{nj}\}$ is invariant to the row-ordering $i$, the column identities $\mathbf{c}_j$ are invariant to $\mathbf{\Pi}_r$. However, permuting the columns by $\mathbf{\Pi}_c$ reorders the resulting set of identities. Thus, the matrix of identities $\mathbf{C} \in \mathbb{R}^{m \times d}$ is transformed as follows: $\mathbf{C}' = \mathbf{\Pi}_c^{\top} \mathbf{C}$.

For \textbf{Component 2 (inter-feature attention)}, by Lemma \ref{lemma:transformer}, the refined identities $\hat{\mathbf{C}}$ remain column-equivariant: $\hat{\mathbf{C}}' = \mathbf{\Pi}_c^{\top} \hat{\mathbf{C}}$.

For \textbf{Component 3 (value-identity fusion and row aggregation)}, the cell embeddings are defined as $\mathbf{e}_{ij} = (\mathbf{w}_v x_{ij} + \mathbf{b}_v) + \hat{\mathbf{c}}_j$. Under both permutations, the transformed cell embedding becomes:
\begin{equation}
\mathbf{e}'_{ij} = \mathbf{e}_{\sigma_r(i), \sigma_c(j)},
\end{equation}
where $\sigma_r$ and $\sigma_c$ are the permutations associated with $\mathbf{\Pi}_r$ and $\mathbf{\Pi}_c$. The row embedding $\mathbf{r}_i$ is then generated by mean pooling, which is a symmetric function over the set $\mathcal{F}_i$; thus, it is invariant to the column permutation $\sigma_c$. Consequently, the row embedding vector $\mathbf{r}_i$ is invariant to $\mathbf{\Pi}_c$. The resulting sequence of row embeddings $\mathbf{R} \in \mathbb{R}^{n \times d}$ satisfies:
\begin{equation}
    \mathbf{R}(\mathbf{\Pi}_r \mathbf{X} \mathbf{\Pi}_c) = \mathbf{\Pi}_r \mathbf{R}(\mathbf{X}).
\end{equation}

\paragraph{Phase 2: Dataset-level structural representation.}
In \textbf{Component 4 (row set interaction)}, the row transformer processes $\mathbf{R}$. By Lemma \ref{lemma:transformer}, the interaction is row-equivariant, yielding: $\hat{\mathbf{R}}' = \mathbf{\Pi}_r \hat{\mathbf{R}}$.

Finally, in \textbf{Component 5 (Pooling by Multihead Attention)}, the global summary $\mathbf{z}$ is computed using a learnable seed vector $\mathbf{s}$. As the attention mechanism computes a weighted sum over the input tokens (rows) based on their similarity to the query $\mathbf{s}$, the output is invariant to the ordering of the input rows \cite{lee2019set}. Therefore, $\text{PMA}(\mathbf{\Pi}_r \hat{\mathbf{R}}) = \text{PMA}(\hat{\mathbf{R}})$.

Combining these steps, we have $f_\theta(\mathbf{\Pi}_r \mathbf{X} \mathbf{\Pi}_c) = f_\theta(\mathbf{X})$, which completes the proof.
\end{proof}

To empirically validate these theoretical guarantees, we evaluated the model across 500 randomized permutations of rows and columns. Across all trials, the predicted number of clusters remained identical, with maximum absolute differences in output logits below $10^{-6}$.

\subsection{Model configuration} \label{app:model_config}
The selected TabK architecture uses an embedding dimension of $d = 256$ with $h = 8$ attention heads. The QFE encodes 50 evenly spaced quantiles per column, processed by a two-layer MLP to produce column identity vectors $\mathbf{c}_j \in \mathbb{R}^d$. A transformer encoder with a feed-forward dimension of $4d$ refines these identities via inter-column self-attention. The row-level transformer encoder uses 4 layers with identical hyperparameters. Dropout is set to 0.3 throughout the backbone. PMA pooling uses $h = 8$ heads, followed by LayerNorm. The prediction head applies a LayerNorm to $\mathbf{z}$, followed by a two-layer MLP with a GELU activation and an intermediate dropout of 0.2, mapping to a distribution over the 14 candidate values of $k \in \{2, \ldots, 15\}$. The logits are passed through log-softmax, and the training loss is the DLDL objective $\mathcal{L} = \mathrm{KL}(q_{k^*} \,\|\, \hat{p})$ of Section~\ref{subsec:amortized_inference}, with $\sigma = 0.5$. The full TabK model contains 4.4M trainable parameters.

We additionally evaluated four alternative prediction heads in the ablation study (see Table \ref{tab:ablations}). The classification head applies standard cross-entropy loss over one-hot targets. The Focal Loss head uses cross-entropy with $(1-p_y)^\gamma$ modulation \cite{lin2017focal}. The ordinal head employs a cumulative threshold model, where each of the 13 binary logits predicts $P(k > t)$. Finally, the regression head produces a single scalar output trained with Huber loss ($\delta = 1.0$). For the classification and focal heads, predictions are decoded via argmax over raw logits. For the regression head, the scalar output is rounded and clamped to the valid range. For the ordinal head, predictions are obtained by counting the number of cumulative probabilities exceeding 0.5 and adding $k_{\min}=2$. All heads share the same MLP backbone, differing only in output dimensionality and loss function.

\subsection{Training and inference protocol} \label{app:training_protocol}
Training uses 5-fold cross-validation on the training split of the synthetic prior. Each fold trains a fresh model for up to 20 epochs with early stopping (patience of 5 epochs, monitored on validation MAE). The best checkpoint per fold is restored after training completes. We use AdamW with a learning rate of $2 \times 10^{-4}$ and a weight decay of $10^{-3}$. Weight decay is applied exclusively to parameters with dimensionality greater than 1 (i.e., weight matrices); biases and LayerNorm scales are exempt. Gradient accumulation over 8 micro-batches of 16 samples yields an effective batch size of 128. A two-phase schedule is employed for the learning rate. A linear warmup from $1\%$ of the base learning rate over $\max(100, 0.05 \times T)$ steps, where $T$ is the total number of optimizer steps, is followed by cosine annealing to $10\%$ of the base rate over the remaining steps. These two phases are composed via a sequential scheduler. To mitigate the quadratic time and memory complexity of the attention mechanism, training uses automatic mixed-precision (AMP) with BFloat16 on Ampere or newer GPUs, while GPU inference uses AMP with Float16. At inference time, the checkpoints are loaded as an ensemble. Each loaded model independently produces a log-probability vector in a single forward pass. The 5 output vectors are averaged in log-space and renormalized, and the final prediction $\bar{k}$ is the argmax of the averaged output. No explicit calibration or temperature scaling is applied.

\subsection{Computational resources} \label{app:compute}
Training was conducted on a single NVIDIA A100-SXM4-40GB GPU using Python 3.13 and PyTorch 2.9.1+cu126. The final 5-fold training run completed in 135,262 seconds ($\approx$37.6 hours). Following the methodology of \href{https://mlco2.github.io/impact}{ML CO2 Impact}, we estimate the carbon footprint as follows. The A100-SXM4-40GB has a thermal design power of 400W. Assuming a power usage effectiveness (PUE) of 1.1 and the average carbon intensity of 0.39 kg CO$_2$/kWh, the final model consumed approximately 16.5 kWh ($\approx$6.4 kg CO$_2$).

\subsection{Hyperparameter selection} \label{app:hpo}
Hyperparameter optimization was conducted via grid search (see Table \ref{tab:hpo_space}), with model selection determined by validation MAE across a 5-fold cross-validation scheme. The quantile resolution of the QFE directly affects the model's structural representation capacity. Empirically, 50 quantile bins achieve the best performance-efficiency trade-off, suggesting that 25 bins may be insufficient to capture fine-grained distributional structure, and 100 bins introduce computational overhead without improving accuracy. Similarly, the DLDL bandwidth $\sigma = 0.5$ consistently outperforms both narrower ($\sigma = 0.1$) and wider ($\sigma = 1.0$) targets. This suggests that a moderate degree of ordinal smoothing provides the ideal bias-variance trade-off, proportionally penalizing predictions without excessively flattening the target distribution.

\begin{table}[t]
  \caption{Hyperparameter search space. Bold values denote the selected configuration for the final TabK model.}
  \label{tab:hpo_space}
  \centering
  \small
  \begin{tabular}{l c l}
    \toprule
    \textbf{Hyperparameter} & \textbf{Symbol} & \textbf{Values} \\
    \midrule
    DLDL bandwidth & $\sigma$ & 0.1, \quad 0.25, \quad $\mathbf{0.5}$, \quad 1.0 \\
    Learning rate & $\eta$ & $10^{-4}$, \quad $\mathbf{2 \times 10^{-4}}$, \quad $5 \times 10^{-4}$ \\
    Embedding dimension & $d_{\text{model}}$ & 128, \quad $\mathbf{256}$, \quad 512 \\
    QFE quantile bins & $b$ & 25, \quad $\mathbf{50}$, \quad 100 \\
    Dropout rate & $p_{\text{drop}}$ & 0.1, \quad $\mathbf{0.3}$, \quad 0.5 \\
    Attention heads & $n_{\text{head}}$ & 4, \quad $\mathbf{8}$, \quad 16 \\
    Transformer layers & $n_{\text{layers}}$ & 2, \quad $\mathbf{4}$, \quad 6 \\
    Weight decay & $\lambda$ & $\mathbf{10^{-3}}$, \quad $5 \times 10^{-3}$ \\
    \bottomrule
  \end{tabular}
\end{table}

\section{Benchmark dataset specifications and profiles} \label{app:data_preprocessing}
To rigorously evaluate TabK's generalization capacity across heterogeneous tabular geometries, we constructed a comprehensive benchmark suite through a systematic selection process from the OpenML database \cite{bischl2021openml}. To align the evaluation with our target operational window while explicitly testing the model's extrapolation capacity beyond the training object-count ceiling, datasets were restricted to classification tasks with a maximum of 10,000 objects, 200 features, and 15 classes (which serve as the ground-truth number of clusters, $k^*$). We subsequently removed duplicate entries and excluded all datasets used in the structural parameter analysis (see Appendix \ref{app:config_sampling}) to prevent distributional leakage into the configuration prior, yielding a final evaluation suite of 50 datasets. The complete specifications for the benchmark suite are summarized in Table \ref{tab:datasets}.

\begin{table}[p]
\caption{Summary of the 50 real-world benchmarks used in our evaluation. OpenML dataset IDs are provided for reproducibility. $n$: number of objects, $m$: number of features, $k$: ground-truth number of clusters.}
\label{tab:datasets}
\small
\begin{tabular}{rlrrrr r}
\toprule
\textbf{ID} & \textbf{Name} & \textbf{Objects ($n$)} &\textbf{ Features ($m$)} & \textbf{$k$} & \textbf{\#Missing} & \textbf{ \#Categorical} \\
\midrule
2 & anneal & 898 & 38 & 5 & 22175 & 32 \\
3 & kr-vs-kp & 3196 & 36 & 2 & 0 & 36 \\
4 & labor & 57 & 16 & 2 & 326 & 8 \\
10 & lymph & 148 & 18 & 4 & 0 & 15 \\
14 & mfeat-fourier & 2000 & 76 & 10 & 0 & 0 \\
15 & breast-w & 699 & 9 & 2 & 16 & 0 \\
22 & mfeat-zernike & 2000 & 47 & 10 & 0 & 0 \\
28 & optdigits & 5620 & 64 & 10 & 0 & 0 \\
29 & credit-a & 690 & 15 & 2 & 67 & 9 \\
31 & credit-g & 1000 & 20 & 2 & 0 & 13 \\
36 & segment & 2310 & 19 & 7 & 0 & 0 \\
43 & haberman & 306 & 3 & 2 & 0 & 1 \\
44 & spambase & 4601 & 57 & 2 & 0 & 0 \\
50 & tic-tac-toe & 958 & 9 & 2 & 0 & 9 \\
54 & vehicle & 846 & 18 & 4 & 0 & 0 \\
55 & hepatitis & 155 & 19 & 2 & 167 & 13 \\
56 & vote & 435 & 16 & 2 & 392 & 16 \\
59 & ionosphere & 351 & 34 & 2 & 0 & 0 \\
61 & iris & 150 & 4 & 3 & 0 & 0 \\
182 & satimage & 6430 & 36 & 6 & 0 & 0 \\
187 & wine & 178 & 13 & 3 & 0 & 0 \\
299 & libras\_move & 360 & 90 & 15 & 0 & 0 \\
694 & diggle\_table\_a2 & 310 & 8 & 9 & 0 & 0 \\
721 & pwLinear & 200 & 10 & 2 & 0 & 0 \\
733 & machine\_cpu & 209 & 6 & 2 & 0 & 0 \\
960 & postoperative-patient-data & 90 & 8 & 2 & 3 & 8 \\
1049 & pc4 & 1458 & 37 & 2 & 0 & 0 \\
1050 & pc3 & 1563 & 37 & 2 & 0 & 0 \\
1063 & kc2 & 522 & 21 & 2 & 0 & 0 \\
1067 & kc1 & 2109 & 21 & 2 & 0 & 0 \\
1068 & pc1 & 1109 & 21 & 2 & 0 & 0 \\
1462 & banknote-authentication & 1372 & 4 & 2 & 0 & 0 \\
1464 & blood-transfusion & 748 & 4 & 2 & 0 & 0 \\
1465 & breast-tissue & 106 & 9 & 6 & 0 & 0 \\
1488 & parkinsons & 195 & 22 & 2 & 0 & 0 \\
1494 & qsar-biodeg & 1055 & 41 & 2 & 0 & 0 \\
1499 & seeds & 210 & 7 & 3 & 0 & 0 \\
1510 & wdbc & 569 & 30 & 2 & 0 & 0 \\
1523 & vertebra-column & 310 & 6 & 3 & 0 & 0 \\
1600 & SPECTF & 267 & 44 & 2 & 0 & 0 \\
4153 & Smartphone & 180 & 67 & 6 & 0 & 1 \\
6332 & cylinder-bands & 540 & 39 & 2 & 999 & 21 \\
40682 & thyroid-new & 215 & 5 & 3 & 0 & 0 \\
40691 & wine-quality-red & 1599 & 11 & 6 & 0 & 0 \\
40701 & churn & 5000 & 20 & 2 & 0 & 4 \\
40975 & car & 1728 & 6 & 4 & 0 & 6 \\
40983 & wilt & 4839 & 5 & 2 & 0 & 0 \\
43922 & mushroom & 8124 & 22 & 2 & 0 & 22 \\
46592 & cervical-cancer & 858 & 32 & 5 & 3622 & 0 \\
47154 & vowel & 990 & 10 & 11 & 0 & 0 \\
\bottomrule
\end{tabular}
\end{table}

\section{Preprocessing pipeline} \label{app:preprocessing}
To ensure a fair comparison across all evaluated baselines, many of which lack native mechanisms for handling incomplete or non-numeric data, we applied a standard preprocessing pipeline to all datasets. Categorical variables were transformed via one-hot encoding (OHE), missing values were replaced using median imputation, and features were standardized to zero mean and unit variance. While deep tabular networks frequently require specialized embedding tables or explicit missingness indicators to process these artifacts without degradation, TabK captures the statistical signatures introduced by standard preprocessing through its Quantile Feature Encoder (QFE). For an OHE column, the sorted quantiles manifest as a distinct two-level step function, explicitly encoding the empirical category frequency as the column's structural identity. Similarly, median imputation introduces an artificial point mass at the median value. In the QFE, this manifests as a flat plateau within the sampled quantiles. By passing this distributional anomaly through the statistical encoder, TabK enables the subsequent attention layers to identify distributional signatures of heavily imputed features. This reduces reliance on explicit missingness indicators while preserving the structural signal introduced by imputation. We evaluated the model's performance in our stratified performance analysis as shown in Figure \ref{fig:stratification_characteristics}, where TabK maintains a consistently low error profile even on datasets with missing values and categorical features.

\section{Extended experimental results} \label{app:extended_exp}
\subsection{Performance on synthetic holdout sets}
While our real-world benchmarks establish that TabK generalizes beyond its training distribution, we evaluated its performance on a synthetic holdout set of 20,000 datasets (a 10\% split of the total generated prior) to isolate its performance across specific geometric topologies. As shown in Table \ref{tab:synthetic_holdout}, TabK achieves an overall MAE of 1.56 and an Exact Match rate of 45.5\%. These metrics are consistent with, though slightly below, the real-world OpenML results (MAE 1.04, 50.0\% Exact Match). We attribute this difference in part to the distribution shift of $k$: the synthetic holdout samples $k$ from a Zipf prior over [2, 15], whereas the real-world benchmarks are skewed toward smaller $k^*$, producing a less challenging marginal distribution for inference. Across synthetic datasets used in our experiments, performance degrades most on the \citet{zellinger2025natural} data (MAE 2.05) that encompasses non-linear manifold transformations warping standard anisotropic overlaps into highly complex, non-convex boundaries. Crucially, the $\pm1$ Tolerance ($\tau \leq 1$) column shows that predictions fall within $\pm 1$ of the true $k$ in 61.8–73.6\% of cases across all geometric domains.

\begin{table}[t]
    \caption{TabK performance on the 10\% synthetic holdout set, stratified by generative strategy. The model demonstrates competitive exact recovery and $\pm1$ Tolerance ($\tau \leq 1$) across diverse geometric domains.}
  \label{tab:synthetic_holdout}
  \centering
  \small
  \begin{tabular}{l c c c}
    \toprule
    \textbf{Generative strategy} & \textbf{MAE $\downarrow$} & \textbf{Exact Match (\%) $\uparrow$} & \textbf{$\tau \le 1$ (\%) $\uparrow$} \\
    \midrule
    Concentric Hyperspheres & 1.17 & 47.3 & 73.6 \\
    \citet{rodriguez2019clustering} & 1.27 & 52.3 & 73.6 \\
    \citet{fachada2023generating} & 1.56 & 41.5 & 66.3 \\
    \citet{jahn2024densired} & 1.76 & 43.8 & 65.3 \\
    \citet{zellinger2025natural} & 2.05 & 42.3 & 61.8 \\
    \midrule
    \textbf{Overall synthetic holdout} & \textbf{1.56} & \textbf{45.5} & \textbf{68.2} \\
    \bottomrule
  \end{tabular}
\end{table}

\subsection{Stratified performance analysis}
\begin{figure}[t]
\centering
\includegraphics[width=\linewidth]{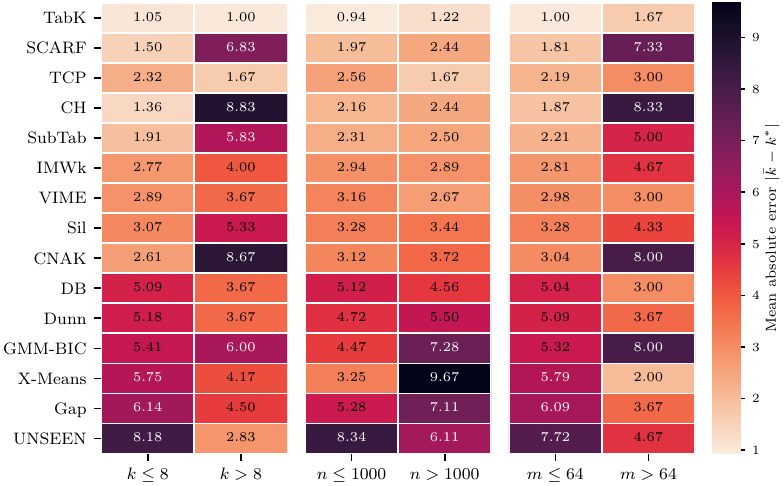}
\caption{ Structural sensitivity analysis. MAE stratified by ground-truth number of clusters ($k$), objects ($n$), and features ($m$). TabK maintains consistent performance across all structural regimes.}
\label{fig:stratification_structure}
\end{figure}

Figure \ref{fig:stratification_structure} presents the results of structural sensitivity analysis in which MAE is stratified by the number of clusters, objects, and features. TabK maintains a consistent error profile across all three dimensions, achieving the MAE values of 1.05 and 1.00 for low and high numbers of clusters, 0.94 and 1.22 for small and large object counts, and 1.00 and 1.67 for low and high feature counts. The modest degradation in high-feature regimes is anticipated, as high-dimensional spaces attenuate pairwise separability signals; however, TabK's QFE mitigates this by operating on marginal quantile representations rather than pairwise distances, retaining the best high-feature MAE among all evaluated methods. TabClustPFN is most sensitive to object count, degrading to an MAE of 2.56 on small datasets, likely because its assignment-based cardinality predictor requires a sufficient object density to form stable cluster assignments.

\begin{figure}[t]
\centering
\includegraphics[width=\linewidth]{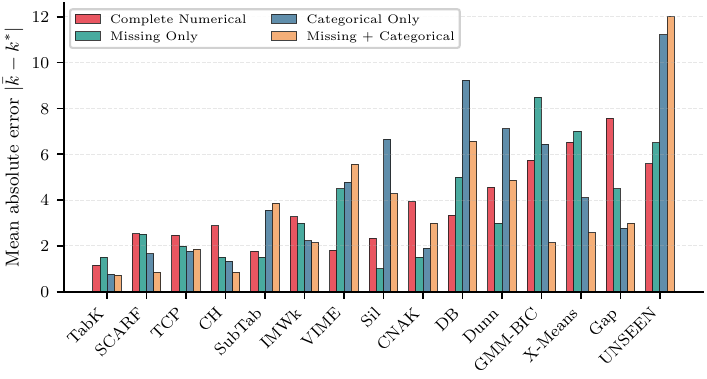}
\caption{MAE stratified by the missing values and categorical features. TabK exhibits a bounded and consistent error profile across all evaluated domains compared to baseline methods and state-of-the-art TabClustPFN.}
\label{fig:stratification_characteristics}
\end{figure}

In Figure \ref{fig:stratification_characteristics}, MAE is stratified by dataset feature composition across four distinct regimes: complete numerical data, numerical data with missing values, data containing categorical features (without missing values), and the highly challenging mixed regime containing both. TabK maintains a consistently low error profile across all four settings. On categorical inputs, one-hot encoding produces characteristic sparsity patterns that the QFE detects via point masses in the empirical distribution. Consequently, even in the hardest mixed regime, where both sparsity patterns and imputation plateaus co-occur, TabK successfully encodes these co-occurring structural signatures and maintains a low error level.

\subsection{Prior-constrained domain evaluation}\label{app:hometurf}
A potential confounding factor when comparing PFNs is the scale of their respective synthetic training distributions. TabClustPFN is trained on a restricted prior bounded by $n \leq 1000$, $m \leq 64$, and $k \leq 10$. To verify that TabK's advantage reported in Table \ref{tab:main_results} persists within TabClustPFN's training-scale upper bounds, we isolated the 27 OpenML datasets in our benchmark that satisfy these upper bounds for a direct head-to-head comparison. As shown in Table \ref{tab:tcp_prior_results}, TabK outperforms TabClustPFN across all metrics even within this restricted sub-domain. For instance, TabK reduces MAE from 2.56 to 0.78 (a 70\% reduction) and achieves an Exact Match rate of 55.6\%, compared to 29.6\% for TabClustPFN. These results indicate that the performance gap is not explained by evaluation beyond TabClustPFN's training-scale upper bounds.

\begin{table}[t]
  \caption{Head-to-head evaluation restricted to TabClustPFN's training-scale upper bounds ($n \leq 1000$, $m \leq 64$, $k \leq 10$), comprising 27 OpenML benchmarks.}
  \label{tab:tcp_prior_results}
  \centering
  \begin{tabular}{l c c c c c}
    \toprule
    Method &  MAE $\downarrow$ & ARI ($k$-means) $\uparrow$ & EM (\%) $\uparrow$ & $\tau\leq1$ (\%) $\uparrow$ & Latency (s) $\downarrow$ \\
    \midrule
    \textbf{TabK} & \textbf{0.78} & \textbf{0.269} & \textbf{55.6} & \textbf{85.2} & \textbf{0.08} \\
    TabClustPFN & 2.56 & 0.235 & 29.6 & 59.3 & 0.50 \\
    \bottomrule
  \end{tabular}
\end{table}

\subsection{Stability under uniform subsampling}\label{app:subsampling}
In Appendix \ref{app:limitation}, we noted that TabK's reliance on full-matrix self-attention imposes an $\mathcal{O}(n^2)$ time and memory complexity. However, because the number of clusters, $k$, is inherently a global property rather than an instance-level artifact, the model does not require observing every object to infer the latent structure. To validate this, we evaluated TabK's prediction consistency under uniform row subsampling on the subset of benchmark datasets exceeding the training prior ceiling ($n>2,500$), across 15 independent draws per sample size. As detailed in Table \ref{tab:subsampling}, TabK exhibits high stability under moderate-to-aggressive subsampling rates. When uniformly subsampling the input matrix to $N=2500$ rows, TabK recovers the same $k$ as the full dataset in 94.3\% of trials. This empirical robustness demonstrates that practitioners can apply uniform row subsampling to mitigate the $\mathcal{O}(n^2)$ attention bottleneck with only modest reductions in prediction consistency, provided a sufficient sample size is retained.

\begin{table}[t]
  \caption{Evaluation of TabK's structural inference stability under uniform row subsampling. Results are aggregated over 15 independent trials for each sample size. The metrics demonstrate that TabK structural inference remains stable across a wide range of sample sizes.}
  \label{tab:subsampling}
  \centering
  \begin{tabular}{l c c c}
    \toprule
    Sample size & Exact Match rate vs. Full & Mean $|\Delta k|$ vs. Full & Latency (ms) \\
    \midrule
    $N=2500$ & 94.3\% & 0.16  & 33.91 \\
    $N=2000$ & 89.5\% & 0.18  & 25.83 \\
    $N=1000$ & 90.5\% & 0.34  & 20.44 \\
    $N=500$  & 86.7\% & 0.34  & 20.96 \\
    $N=250$  & 75.2\% & 0.62  & 19.27 \\
    $N=100$  & 61.9\% & 0.73  & 19.20 \\
    \bottomrule
  \end{tabular}
\end{table}

\subsection{Robustness to noisy features and outliers}\label{app:robustness}
We evaluated TabK on all 50 real-world benchmarks under two perturbations that leave the annotated $k^*$ unchanged, averaging each nonzero level over five seeds. For noisy features, we append independent standard Gaussian columns equal to 25\%, 50\%, or 100\% of the original feature count. For diffuse outliers, we append points equal to 1\%, 5\%, or 15\% of the sample size, drawn independently per feature from the observed range extended by 25\% on either side. Each perturbed dataset is then standardized. Table~\ref{tab:robustness} reports the MAE and the agreement with TabK's prediction on the clean data. Even when the feature count is doubled with uninformative columns, TabK's MAE (1.94) remains below the clean-data MAE of every competing method in Table~\ref{tab:main_results}.

\begin{table}[t]
\caption{TabK under noisy features and diffuse outliers on the 50 benchmarks (mean over five seeds). Agreement is the percentage of datasets on which the prediction matches the prediction on clean data.}
\label{tab:robustness}
\centering
\small
\begin{tabular}{l l c c}
\toprule
Perturbation & Level & MAE $\downarrow$ & Agreement (\%) $\uparrow$ \\
\midrule
None (clean) & -- & 1.04 & 100.0 \\
\midrule
Noisy features & +25\% of $m$ & 1.43 & 86.8 \\
 & +50\% of $m$ & 1.56 & 85.6 \\
 & +100\% of $m$ & 1.94 & 76.8 \\
\midrule
Diffuse outliers & 1\% of $n$ & 1.30 & 80.4 \\
 & 5\% of $n$ & 1.43 & 68.0 \\
 & 15\% of $n$ & 1.61 & 67.6 \\
\bottomrule
\end{tabular}
\end{table}

\subsection{Evaluation beyond the training scale}\label{app:scale_external}
To evaluate TabK beyond the primary benchmark, we assembled 25 datasets from the OpenML-CC18 suite \cite{bischl2021openml}, after removing every dataset in the primary benchmark and every dataset used to estimate the prior's $(k,n,m)$ marginals. The suite spans 500 to 96{,}320 objects and 5 to 1{,}776 features; the 10 datasets above 500{,}000 cells exceed the largest synthetic training table ($2{,}500 \times 200$) by up to $13\times$. Each dataset is processed at its full size, without subsampling or truncation. Table~\ref{tab:scale_external} summarizes the results and Table~\ref{tab:cc18_per_dataset} lists the per-dataset predictions. On both subsets, TabK improves on TabClustPFN in every reported metric.

\begin{table}[t]
\caption{Evaluation beyond the primary benchmark. OpenML-CC18: 25 datasets disjoint from the primary benchmark and from the datasets used to estimate the prior's $(k,n,m)$ marginals; the $>$500k subset exceeds the largest synthetic training table.}
\label{tab:scale_external}
\centering
\small
\setlength{\tabcolsep}{4pt}
\begin{tabular}{l l c c c c c c}
\toprule
Subset & Method & MAE $\downarrow$ & EM (\%) $\uparrow$ & $\tau \leq 1$ (\%) $\uparrow$ & Brier $\downarrow$ & NLL $\downarrow$ & Latency (s) $\downarrow$ \\
\midrule
CC18, all (25) & \textbf{TabK} & \textbf{1.56} & \textbf{44.0} & \textbf{60.0} & \textbf{0.718} & \textbf{2.106} & \textbf{0.70} \\
 & TCP & 2.60 & 40.0 & 56.0 & 0.897 & 5.274 & 4.31 \\
\midrule
CC18, $>$500k cells (10) & \textbf{TabK} & \textbf{1.00} & \textbf{60.0} & \textbf{80.0} & \textbf{0.542} & \textbf{1.238} & \textbf{1.66} \\
 & TCP & 2.00 & 50.0 & 70.0 & 0.904 & 6.180 & 9.88 \\
\bottomrule
\end{tabular}
\end{table}

\begin{table}[t]
\caption{Per-dataset predictions $\bar{k}$ on the 25 OpenML-CC18 datasets used for the scale evaluation, ordered by table size. Bold entries indicate exact recovery; underlined entries indicate $|\bar{k}-k^*|=1$.}
\label{tab:cc18_per_dataset}
\centering
\small
\begin{tabular}{r l r r r c c c}
\toprule
ID & Dataset & $n$ & $m$ & Cells & $k^*$ & TabK & TCP \\
\midrule
1480 & ilpd & 583 & 12 & 6{,}996 & 2 & \textbf{2} & \textbf{2} \\
469 & analcatdata\_dmft & 797 & 9 & 7{,}173 & 6 & 2 & 10 \\
40994 & climate-model-simulation-crashes & 540 & 18 & 9{,}720 & 2 & \textbf{2} & \textbf{2} \\
23 & cmc & 1{,}473 & 9 & 13{,}257 & 3 & 7 & \underline{2} \\
1489 & phoneme & 5{,}404 & 5 & 27{,}020 & 2 & 4 & 9 \\
40982 & steel-plates-fault & 1{,}941 & 27 & 52{,}407 & 7 & 2 & 10 \\
188 & eucalyptus & 736 & 96 & 70{,}656 & 5 & 2 & 2 \\
23381 & dresses-sales & 500 & 167 & 83{,}500 & 2 & \textbf{2} & \textbf{2} \\
1497 & wall-robot-navigation & 5{,}456 & 24 & 130{,}944 & 4 & 8 & 10 \\
1487 & ozone-level-8hr & 2{,}534 & 72 & 182{,}448 & 2 & 5 & 8 \\
40499 & texture & 5{,}500 & 40 & 220{,}000 & 11 & 9 & 2 \\
1053 & jm1 & 10{,}885 & 21 & 228{,}585 & 2 & \textbf{2} & \underline{3} \\
38 & sick & 3{,}772 & 74 & 279{,}128 & 2 & \textbf{2} & \textbf{2} \\
4534 & PhishingWebsites & 11{,}055 & 30 & 331{,}650 & 2 & \underline{3} & \textbf{2} \\
151 & electricity & 45{,}312 & 8 & 362{,}496 & 2 & \underline{3} & 7 \\
\midrule
40670 & dna & 3{,}186 & 180 & 573{,}480 & 3 & \textbf{3} & \underline{2} \\
1468 & cnae-9 & 1{,}080 & 856 & 924{,}480 & 9 & 14 & 2 \\
46 & splice & 3{,}190 & 347 & 1{,}106{,}930 & 3 & \underline{2} & \underline{2} \\
23517 & numerai28.6 & 96{,}320 & 21 & 2{,}022{,}720 & 2 & \textbf{2} & 10 \\
1461 & bank-marketing & 45{,}211 & 60 & 2{,}712{,}660 & 2 & \textbf{2} & \textbf{2} \\
1486 & nomao & 34{,}465 & 118 & 4{,}066{,}870 & 2 & \underline{3} & \textbf{2} \\
40978 & Internet-Advertisements & 3{,}279 & 1{,}558 & 5{,}108{,}682 & 2 & \textbf{2} & \textbf{2} \\
1590 & adult & 48{,}842 & 113 & 5{,}519{,}146 & 2 & \textbf{2} & \textbf{2} \\
1478 & har & 10{,}299 & 561 & 5{,}777{,}739 & 6 & 3 & 3 \\
4134 & Bioresponse & 3{,}751 & 1{,}776 & 6{,}661{,}776 & 2 & \textbf{2} & \textbf{2} \\
\bottomrule
\end{tabular}
\end{table}

\subsection{Failure analysis}\label{app:failure_analysis}
While TabK consistently improves structural recovery compared to evaluated methods, an analysis of its failure modes provides valuable insight into the inherent ambiguity of tabular clustering. By examining the datasets where TabK registers its highest absolute errors, we observe that these failures in part stem from a misalignment between human-annotated class labels and the intrinsic spatial geometry of the data (see Appendix \ref{app:limitation}). We categorize these failure instances into two distinct phenomena.

\paragraph{Hierarchical ambiguity.} In datasets with highly granular human annotations, the most mathematically robust geometric partitions often do not align with the fine-grained label structure. For example, on the \textit{Smartphone} dataset (\texttt{ID=4153}), the annotated number of clusters is $k^*=6$ (representing specific movements like walking, sitting, and standing). TabK assigns its highest confidence (51.9\%) to $k=2$. While evaluated as an error of 4, an inspection of the baseline algorithms reveals that the Calinski--Harabasz index, the Davies--Bouldin index, the Silhouette width, and the deep baseline TabClustPFN model all unanimously predict $k=2$. This indicates that the global geometry is dominated by two highly separable macro-clusters (e.g., active versus passive movements). Consistent with these baselines, TabK identifies the dominant macro-level partition rather than fine-grained sub-class boundaries.

\paragraph{Unresolvable geometries and flat posteriors.} The second category of error occurs in datasets that appear to lack clear spatial separability. For instance, on the \textit{vehicle} (\texttt{ID=54}, $k^*=4$, predicted $\bar{k}=8$) and \textit{wine-quality-red} (\texttt{ID=40691}, $k^*=6$, predicted $\bar{k}=2$) datasets, TabK fails to recover the ground truth. However, the DLDL predictive head provides a crucial diagnostic advantage over classical heuristics: it outputs a probability mass function. For the \textit{vehicle} dataset, TabK's maximum predicted confidence is merely 10.7\%, whereas it is 12.1\% for the \textit{wine-quality-red} dataset, with the probability mass distributed almost uniformly across $k$. Rather than generating an overconfident exact prediction, TabK's output distribution reflects the uncertainty of these estimates.

\subsection{Full per-dataset results}\label{app:full_results}
To ensure complete empirical transparency, Table \ref{tab:per_dataset_k} details the exact numbers of clusters, $\bar{k}$, predicted by TabK and all other evaluated methods across the 50 OpenML benchmarks considered.

\begin{table}[p]
\caption{Predicted numbers of clusters, $\bar{k}$, reported for the 50 OpenML benchmarks. OpenML dataset IDs are indicated. Methods are ordered by the MAE performance (best first). Bold entries indicate exact recovery ($\bar{k} = k^*$). Underlined entries indicate a prediction with $\tau \leq 1$ ($|\bar{k} - k^*| = 1$).}
\label{tab:per_dataset_k}
\centering
\footnotesize
\setlength{\tabcolsep}{2.5pt}
\begin{tabular}{l c c c c c c c c c c c c c c c c c c c }
\toprule
ID & $k^*$ & \rotatebox{90}{TabK} & \rotatebox{90}{SCARF+CH} & \rotatebox{90}{TCP} & \rotatebox{90}{CH} & \rotatebox{90}{SubTab+Sil} & \rotatebox{90}{SCARF+Sil} & \rotatebox{90}{IMWk} & \rotatebox{90}{VIME+Sil} & \rotatebox{90}{VIME+CH} & \rotatebox{90}{Sil} & \rotatebox{90}{CNAK} & \rotatebox{90}{SubTab+CH} & \rotatebox{90}{DB} & \rotatebox{90}{Dunn} & \rotatebox{90}{GMM-BIC} & \rotatebox{90}{X-Means} & \rotatebox{90}{Gap} & \rotatebox{90}{UNSEEN} \\
\midrule
2 & 5 & 2 & \textbf{5} & 2 & 3 & 15 & 3 & \underline{4} & 15 & 2 & 8 & 2 & 14 & 12 & \underline{4} & \textbf{5} & 15 & 2 & 13 \\
3 & 2 & \textbf{2} & 4 & \textbf{2} & \textbf{2} & 9 & \textbf{2} & 4 & \textbf{2} & \textbf{2} & 7 & \textbf{2} & \underline{3} & 15 & 13 & 5 & 15 & \textbf{2} & 15 \\
4 & 2 & \underline{3} & \textbf{2} & 10 & \textbf{2} & \textbf{2} & \underline{3} & 4 & \textbf{2} & \textbf{2} & 15 & \textbf{2} & \textbf{2} & 15 & 14 & 4 & \textbf{2} & \underline{3} & 15 \\
10 & 4 & \underline{3} & \underline{3} & 2 & 2 & \underline{3} & \textbf{4} & \textbf{4} & 2 & 2 & 2 & 2 & \underline{5} & 14 & 7 & \underline{3} & 2 & 2 & 15 \\
14 & 10 & \textbf{10} & 3 & \textbf{10} & 2 & 2 & 2 & 8 & 15 & 2 & 2 & 3 & 2 & 15 & 3 & 3 & 15 & 15 & 15 \\
15 & 2 & \textbf{2} & 5 & \underline{3} & \textbf{2} & \textbf{2} & \textbf{2} & 7 & \textbf{2} & \underline{3} & \textbf{2} & \textbf{2} & \textbf{2} & \textbf{2} & 5 & 15 & 6 & 8 & 11 \\
22 & 10 & \underline{9} & 4 & \textbf{10} & 2 & 3 & 2 & \underline{9} & 7 & 2 & 7 & 2 & 2 & 14 & 8 & 5 & 15 & 15 & 14 \\
28 & 10 & 13 & 6 & \textbf{10} & 2 & 13 & \textbf{10} & 6 & \underline{9} & 2 & 14 & 2 & 13 & 14 & 13 & \underline{11} & 15 & 2 & 12 \\
29 & 2 & \textbf{2} & \underline{3} & \underline{3} & \underline{3} & \textbf{2} & \textbf{2} & 5 & 5 & \textbf{2} & \textbf{2} & \underline{3} & \underline{3} & \textbf{2} & \textbf{2} & 6 & 4 & 8 & 15 \\
31 & 2 & \textbf{2} & 4 & \textbf{2} & \textbf{2} & \textbf{2} & 6 & 7 & \textbf{2} & \textbf{2} & 15 & \textbf{2} & \textbf{2} & 15 & 12 & 7 & \underline{3} & \textbf{2} & 15 \\
36 & 7 & 2 & 13 & 4 & 2 & \underline{6} & 4 & \textbf{7} & 4 & 2 & \textbf{7} & 2 & \underline{6} & \underline{6} & 14 & 14 & 15 & 2 & 10 \\
43 & 2 & \textbf{2} & 4 & \textbf{2} & 5 & \textbf{2} & 4 & 7 & 15 & 15 & 4 & 5 & 15 & 14 & 12 & 8 & \textbf{2} & 5 & 7 \\
44 & 2 & \textbf{2} & \textbf{2} & \textbf{2} & \textbf{2} & \textbf{2} & \textbf{2} & 6 & \underline{3} & \textbf{2} & \textbf{2} & 4 & \textbf{2} & \textbf{2} & \textbf{2} & 12 & 15 & 15 & 10 \\
50 & 2 & \underline{3} & \textbf{2} & 10 & \textbf{2} & 8 & \textbf{2} & \underline{3} & 13 & \textbf{2} & 14 & 9 & \textbf{2} & 14 & 14 & 14 & \textbf{2} & 12 & 15 \\
54 & 4 & 8 & \textbf{4} & 8 & 2 & 2 & \textbf{4} & 6 & 2 & 2 & 2 & 2 & \textbf{4} & 2 & \textbf{4} & 7 & 15 & 15 & 10 \\
55 & 2 & \textbf{2} & 4 & \textbf{2} & \underline{3} & \textbf{2} & 7 & 6 & 14 & \textbf{2} & \textbf{2} & 12 & \underline{3} & \textbf{2} & \textbf{2} & \textbf{2} & 4 & \underline{3} & 15 \\
56 & 2 & \textbf{2} & \underline{3} & \underline{3} & \textbf{2} & \underline{3} & 9 & \textbf{2} & \textbf{2} & \textbf{2} & \underline{3} & \textbf{2} & \underline{3} & \underline{3} & 13 & 8 & \underline{3} & 8 & 14 \\
59 & 2 & \underline{3} & \textbf{2} & 10 & \textbf{2} & 4 & \textbf{2} & 4 & \underline{3} & \textbf{2} & 4 & \textbf{2} & \textbf{2} & 6 & \textbf{2} & 10 & \underline{3} & 10 & 11 \\
61 & 3 & \textbf{3} & \underline{2} & \underline{4} & \underline{2} & 5 & \underline{2} & 5 & \underline{2} & \underline{2} & \underline{2} & \underline{2} & 15 & \underline{2} & \underline{2} & \underline{2} & 7 & 5 & 15 \\
182 & 6 & 11 & 3 & 2 & 3 & 3 & 2 & \textbf{6} & 3 & 4 & 3 & 2 & 4 & 3 & 15 & \underline{5} & 15 & 15 & 11 \\
187 & 3 & 5 & 7 & \underline{4} & \textbf{3} & \textbf{3} & 7 & 5 & \underline{2} & \underline{2} & \textbf{3} & \textbf{3} & \textbf{3} & \textbf{3} & 13 & 15 & \textbf{3} & \underline{4} & 15 \\
299 & 15 & \underline{14} & 4 & 10 & 2 & 12 & 2 & 6 & \textbf{15} & 2 & \underline{14} & 2 & 5 & \textbf{15} & \textbf{15} & 2 & \textbf{15} & \textbf{15} & \textbf{15} \\
694 & 9 & \textbf{9} & 2 & 5 & 2 & 2 & 2 & 5 & 3 & 15 & 2 & 2 & 15 & 3 & 3 & 15 & 15 & 15 & 14 \\
721 & 2 & \underline{3} & \underline{3} & 10 & \textbf{2} & \textbf{2} & \underline{3} & \textbf{2} & \textbf{2} & \textbf{2} & 14 & \textbf{2} & \textbf{2} & 14 & \textbf{2} & 5 & \textbf{2} & \textbf{2} & 15 \\
733 & 2 & \textbf{2} & 8 & \textbf{2} & \textbf{2} & \textbf{2} & 6 & 6 & \textbf{2} & 15 & \textbf{2} & \textbf{2} & 14 & 12 & 6 & 11 & 4 & 11 & \underline{3} \\
960 & 2 & \underline{3} & \textbf{2} & \textbf{2} & \underline{3} & 15 & 9 & \underline{3} & 15 & \textbf{2} & 15 & 8 & \underline{3} & 15 & 12 & \underline{3} & \textbf{2} & \textbf{2} & 14 \\
1049 & 2 & \underline{3} & 4 & \textbf{2} & \textbf{2} & \textbf{2} & \textbf{2} & 5 & \textbf{2} & \textbf{2} & \textbf{2} & \textbf{2} & \textbf{2} & 10 & 15 & 10 & 15 & 15 & 7 \\
1050 & 2 & \underline{3} & 4 & 4 & \underline{3} & \textbf{2} & 4 & 7 & 7 & \textbf{2} & \textbf{2} & 12 & 9 & \textbf{2} & \textbf{2} & 12 & 15 & 15 & 7 \\
1063 & 2 & \textbf{2} & \underline{3} & \underline{3} & 6 & 4 & \underline{3} & 4 & \underline{3} & 7 & \textbf{2} & 15 & 15 & \textbf{2} & \textbf{2} & 6 & 15 & 15 & 7 \\
1067 & 2 & \textbf{2} & \textbf{2} & 4 & \textbf{2} & \textbf{2} & \textbf{2} & 6 & \textbf{2} & \underline{3} & \textbf{2} & \textbf{2} & 15 & \textbf{2} & \textbf{2} & 12 & 15 & 15 & \textbf{2} \\
1068 & 2 & \textbf{2} & \textbf{2} & \textbf{2} & \underline{3} & \textbf{2} & \textbf{2} & 8 & \underline{3} & \textbf{2} & \textbf{2} & 15 & 8 & \underline{3} & \textbf{2} & 11 & 15 & 15 & 4 \\
1462 & 2 & \underline{3} & \underline{3} & 9 & 5 & \textbf{2} & \textbf{2} & 8 & \underline{3} & 15 & 15 & 6 & 15 & 13 & 14 & 15 & 15 & 7 & 10 \\
1464 & 2 & \textbf{2} & \underline{3} & 10 & 13 & \underline{3} & \textbf{2} & 7 & \underline{3} & 15 & \textbf{2} & 8 & 5 & 14 & 8 & 10 & 13 & 13 & 4 \\
1465 & 6 & 2 & 3 & \textbf{6} & \underline{7} & 3 & 3 & \underline{5} & 3 & 4 & 2 & 2 & 8 & 3 & 3 & 11 & \underline{5} & 10 & 14 \\
1488 & 2 & \textbf{2} & \textbf{2} & 8 & \underline{3} & \underline{3} & \textbf{2} & 6 & \textbf{2} & 4 & \textbf{2} & 5 & \underline{3} & \textbf{2} & 14 & \underline{3} & 6 & 15 & 10 \\
1494 & 2 & \textbf{2} & \textbf{2} & 10 & \textbf{2} & \textbf{2} & \textbf{2} & 6 & 9 & \textbf{2} & \textbf{2} & \textbf{2} & \textbf{2} & 15 & 15 & 12 & 15 & 15 & 9 \\
1499 & 3 & \textbf{3} & \textbf{3} & 5 & \underline{2} & \underline{2} & \underline{4} & 7 & \textbf{3} & \textbf{3} & \underline{2} & \underline{2} & \textbf{3} & \underline{2} & 11 & \underline{4} & \underline{4} & 7 & 15 \\
1510 & 2 & \textbf{2} & \underline{3} & \textbf{2} & \textbf{2} & \textbf{2} & \underline{3} & 6 & \textbf{2} & \textbf{2} & \textbf{2} & \textbf{2} & \textbf{2} & \textbf{2} & 15 & 4 & 10 & 15 & 7 \\
1523 & 3 & \underline{2} & \textbf{3} & \underline{2} & \underline{2} & \underline{2} & \underline{2} & 6 & \textbf{3} & \underline{2} & \underline{2} & \underline{4} & \underline{2} & \underline{4} & 6 & \textbf{3} & \underline{2} & 8 & 7 \\
1600 & 2 & \textbf{2} & \textbf{2} & \textbf{2} & \textbf{2} & \textbf{2} & \textbf{2} & 5 & \textbf{2} & \textbf{2} & \textbf{2} & \textbf{2} & \textbf{2} & \textbf{2} & \textbf{2} & 4 & \underline{3} & 12 & 14 \\
4153 & 6 & 2 & 2 & 2 & 2 & 2 & 2 & 3 & 2 & 2 & 2 & 2 & 2 & 2 & 2 & 2 & \underline{7} & 12 & 15 \\
6332 & 2 & \textbf{2} & 4 & \textbf{2} & \underline{3} & 5 & 4 & 6 & \underline{3} & \textbf{2} & \textbf{2} & \underline{3} & \textbf{2} & 14 & \textbf{2} & 4 & 5 & 6 & 15 \\
40682 & 3 & \underline{2} & \underline{2} & \underline{2} & \textbf{3} & \underline{2} & \underline{2} & 7 & \textbf{3} & 15 & \textbf{3} & \textbf{3} & 6 & 5 & 5 & \underline{4} & \textbf{3} & 8 & 7 \\
40691 & 6 & 2 & 2 & 4 & 2 & 4 & 2 & \underline{7} & 2 & 2 & 2 & 2 & 2 & \textbf{6} & 14 & 9 & 11 & 15 & 10 \\
40701 & 2 & \textbf{2} & \underline{3} & \textbf{2} & \textbf{2} & \textbf{2} & 6 & \textbf{2} & \textbf{2} & \textbf{2} & \textbf{2} & \textbf{2} & \textbf{2} & 7 & \textbf{2} & 7 & 8 & 6 & 15 \\
40975 & 4 & \underline{3} & \underline{5} & 2 & \underline{3} & \underline{3} & 15 & \textbf{4} & \underline{3} & \underline{3} & 15 & \underline{3} & 2 & 15 & 15 & 14 & \underline{3} & \textbf{4} & 15 \\
40983 & 2 & \textbf{2} & 5 & \textbf{2} & \textbf{2} & \textbf{2} & \textbf{2} & 8 & \underline{3} & \underline{3} & \textbf{2} & \underline{3} & \underline{3} & \textbf{2} & \textbf{2} & 9 & 15 & \textbf{2} & 4 \\
43922 & 2 & \textbf{2} & 4 & \textbf{2} & 4 & 15 & \underline{3} & 6 & 14 & 4 & 13 & \textbf{2} & 15 & 5 & 5 & 14 & 15 & \textbf{2} & 15 \\
46592 & 5 & 2 & 3 & 2 & 2 & 2 & 3 & \underline{6} & 14 & \textbf{5} & 3 & 2 & 2 & 15 & 2 & 9 & 15 & 2 & 9 \\
47154 & 11 & \underline{10} & 5 & \underline{10} & 2 & 4 & 5 & 7 & 4 & 4 & 2 & 2 & 4 & 14 & 15 & 15 & 15 & 14 & \underline{12} \\
\midrule
\multicolumn{2}{l}{\textbf{MAE}} & \textbf{1.04} & \underline{2.14} & 2.24 & 2.26 & 2.38 & 2.50 & 2.92 & 2.98 & 3.08 & 3.34 & 3.34 & 3.72 & 4.92 & 5.00 & 5.48 & 5.56 & 5.94 & 7.54 \\
\bottomrule
\end{tabular}
\end{table}

\section{Baseline implementation details}\label{app:baseline}
To ensure a rigorous and reproducible comparison, we provide comprehensive implementation details for the 17 competing methods evaluated in our study (see Table \ref{tab:main_results}). All baselines were evaluated over the same candidate range $k \in \{2, \dots, 15\}$, except TabClustPFN, whose pretrained output range is limited to $k \leq 10$ (see Appendix~\ref{app:hometurf} for a comparison within its domain), using a fixed random seed. 

No method was tuned per dataset or against the real-world annotations: baselines used the configurations recommended by their authors where available, with identical preprocessing. For the four internal cluster validity indices, first, $k$-means (\texttt{scikit-learn}, \texttt{n\_init=100}, \texttt{k-means++} initialization) is executed for every candidate $k$, producing a set of label vectors $\{\hat{\mathbf{y}}_k\}$. Each CVI is then computed on the data--label pairs $(\mathbf{X}, \hat{\mathbf{y}}_k)$. We used the \texttt{scikit-learn} implementations for the Silhouette, Calinski--Harabasz, and Davies--Bouldin indices. We used our own implementation of the Dunn index. Additionally, we implemented the Gap statistic \cite{tibshirani2001gapstat} and GMM-BIC \cite{fraley2002gmmbic} according to their original specifications, and integrated the author-provided MATLAB implementation of IMWk \cite{deamorim2015imwk} into our benchmarking pipeline. All implementations and scripts are available in our repository at \href{https://github.com/mrbakhtyari/TabK}{\nolinkurl{github.com/mrbakhtyari/TabK}}.

Specific details for the remaining methods are as follows:

\begin{itemize}
    \item \textbf{X-Means} \cite{pelleg2000xmeans}: We used the \texttt{PyClustering} implementation with the \texttt{$k$-means++} initialization, and the \texttt{BIC} splitting criterion.
    \item \textbf{CNAK} \cite{saha2021cnak}: We used the official source code at: \href{https://github.com/Jayasree-Saha/CNAK}{Jayasree-Saha/CNAK}.
    \item \textbf{UNSEEN} \cite{leiber2024unseen}: We used the official source code at: \href{https://github.com/collinleiber/UNSEEN}{collinleiber/UNSEEN}.
    \item \textbf{TabClustPFN} \cite{zhao2026tabclustpfn}: We used the official source code at: \href{https://github.com/Tianqi-Zhao/TabClustPFN}{Tianqi-Zhao/TabClustPFN}.
    \item \textbf{VIME, SCARF, and SubTab} \cite{yoon2020vime, bahri2022scarf, ucar2021subtab}: We trained each encoder on the preprocessed dataset and applied the same Silhouette and Calinski--Harabasz selection routine and $k$-means setting (\texttt{n\_init=100}) to the learned embeddings as for the raw-feature CVIs. We used the official implementations of VIME (\href{https://github.com/jsyoon0823/VIME}{jsyoon0823/VIME}) and SubTab (\href{https://github.com/AstraZeneca/SubTab}{AstraZeneca/SubTab}), and a public implementation of SCARF (\href{https://github.com/clabrugere/pytorch-scarf}{clabrugere/pytorch-scarf}).
\end{itemize}

\section{Broader impacts} \label{app:impact}
TabK improves the efficiency of unsupervised learning for tabular data, reducing the computational cost and potentially the energy consumption of standard clustering pipelines. However, automating cluster discovery carries potential societal risks if applied to sensitive human-centric data without oversight. We urge practitioners to manually audit the resulting clusters for fairness and privacy implications before deploying them in sensitive domains.

\section{Asset licenses}  \label{app:licenses}
All datasets and codebases used in this work are publicly available and properly credited; datasets and code are used under the terms of their sources. Identified software licenses include BSD-3-Clause
(scikit-learn, PyClustering, ClustPy, and repliclust), MIT
(PyClugen, DENSIRED, and the SCARF implementation), Apache-2.0
(SubTab), and CC BY-NC 4.0 (TabClustPFN).

\end{document}